\documentclass{article}

\usepackage[nonatbib,final]{neurips_2022}
\usepackage[numbers]{natbib}

\usepackage[utf8]{inputenc} % allow utf-8 input
\usepackage[T1]{fontenc}    % use 8-bit T1 fonts
\usepackage{hyperref}       % hyperlinks
\usepackage{url}            % simple URL typesetting
\usepackage{booktabs}       % professional-quality tables
\usepackage{amsfonts}       % blackboard math symbols
\usepackage{nicefrac}       % compact symbols for 1/2, etc.
\usepackage{microtype}      % microtypography
\usepackage[table]{xcolor}  % colors
\usepackage{enumitem}
\usepackage{graphicx}
\usepackage{array}

\usepackage{changepage,threeparttable} % for wide tables
\usepackage{multirow}
\usepackage{caption}
\usepackage{subcaption}
\usepackage{lipsum}
\usepackage{graphicx}
\usepackage{flushend}
\usepackage{xspace}
\usepackage{wrapfig}
\usepackage{algorithm}

\hypersetup{
    colorlinks,
    linkcolor={red!50!black},
    citecolor={blue!50!black},
    urlcolor={blue!80!black}
}

\graphicspath{{figures/}}

\title{Recipe for a General, Powerful, Scalable\\ Graph Transformer}
\author{%
   Ladislav Rampášek\thanks{To whom correspondence should be addressed: \texttt{ladislav.rampasek@mila.quebec}} \\
   Mila, Université de Montréal \\
   \And
   Mikhail Galkin \\
   Mila, McGill University \\
   \AND
   Vijay Prakash Dwivedi \\
   Nanyang Technological University, Singapore \\
   \And
   Anh Tuan Luu \\
   Nanyang Technological University, Singapore \\
   \And
   Guy Wolf \\
   Mila, Université de Montréal \\
   \And
   Dominique Beaini \\
   Valence Discovery, Mila, Université de Montréal \\
}

\begin{document}

\definecolor{purpleheart}{rgb}{0.41, 0.21, 0.61}
\newcommand{\mycomment}[3]{\textcolor{#1}{[\textbf{#2:} #3]}}
\newcommand{\lr}[1]{\mycomment{orange}{LR}{#1}}
\newcommand{\db}[1]{\mycomment{magenta}{DB}{#1}}
\newcommand{\mg}[1]{\mycomment{green!70!black}{MG}{#1}}
\newcommand{\draft}[1]{\textcolor{blue}{#1}}
\newcommand{\vd}[1]{\textcolor{purpleheart}{#1}} % DB: Sorry, can't distinguish the purple from black, I'm colorblind

\newcommand{\myparagraph}[1]{\noindent\textbf{#1}}

\definecolor{darkgreen}{rgb}{0, 0.7, 0}

\definecolor{dark2green}{rgb}{0.1, 0.65, 0.3}
\definecolor{dark2orange}{rgb}{0.9, 0.4, 0.}
\definecolor{dark2purple}{rgb}{0.4, 0.4, 0.8}
\newcommand{\first}[1]{\textbf{\textcolor{dark2green}{#1}}}
\newcommand{\second}[1]{\textbf{\textcolor{dark2orange}{#1}}}
\newcommand{\third}[1]{\textbf{\textcolor{dark2purple}{#1}}}

\newcommand{\method}{GPS\xspace}
\newcommand{\gtgym}{\textsc{GraphGPS}\xspace}

\definecolor{pe_color}{HTML}{00239C}
\definecolor{se_color}{HTML}{E10600}
\newcommand{\PE}{\textcolor{pe_color}{PE}\xspace}
\newcommand{\SE}{\textcolor{se_color}{SE}\xspace}

\maketitle
\setcounter{footnote}{0}

\begin{abstract}
% We propose a general and scalable foundation for graph Transformers (GT), first by organizing the various ideas around positional and structural encodings, then by open-sourcing a fast and powerful architecture with linear complexity and state-of-the-art results on a diverse set of benchmarks.
We propose a recipe on how to build a general, powerful, scalable (GPS) graph Transformer with linear complexity and state-of-the-art results on a diverse set of benchmarks.
Graph Transformers (GTs) have gained popularity in the field of graph representation learning with a variety of recent publications but they lack a common foundation about what constitutes a good positional or structural encoding, and what differentiates them. In this paper, we summarize the different types of encodings with a clearer definition and categorize them as being \textit{local}, \textit{global} or \textit{relative}.
The prior GTs are constrained to small graphs with a few hundred nodes, here we propose the first architecture with a complexity linear in the number of nodes and edges $O(N+E)$ by decoupling the local real-edge aggregation from the fully-connected Transformer. We argue that this decoupling does not negatively affect the expressivity, with our architecture being a universal function approximator on graphs. Our GPS recipe consists of choosing 3 main ingredients: (i) positional/structural encoding, (ii) local message-passing mechanism, and (iii) global attention mechanism.
We provide a modular framework \textsc{GraphGPS}\footnote{The source code of \textsc{GraphGPS} is available at: \url{https://github.com/rampasek/GraphGPS}.}
that supports multiple types of encodings and that provides efficiency and scalability both in small and large graphs.
We test our architecture on 16 benchmarks and show highly competitive results in all of them, show-casing the empirical benefits gained by the modularity and the combination of different strategies.
\end{abstract}

%##############################################################################
\section{Introduction}
%##############################################################################

Graph Transformers (GTs) alleviate fundamental limitations pertaining to the sparse message passing mechanism, e.g., over-smoothing \cite{Oono2020Graph}, over-squashing \cite{alon2021on}, and expressiveness bounds \cite{xu2018how, morris2019}, by allowing nodes to attend to all other nodes in a graph (\emph{global attention}).
This benefits several real-world applications, such as modeling chemical interactions beyond the covalent bonds \cite{ying2021graphormer}, or graph-based robotic control \cite{kurin2020my_body}.
Global attention, however, requires  nodes to be better identifiable within the graph and its substructures \cite{dwivedi2020generalization}. 
This has led to a flurry of recently proposed fully-connected graph transformer models \cite{dwivedi2020generalization,kreuzer2021rethinking,ying2021graphormer,mialon2021graphit,jain2021graphtrans} as well as various positional encoding schemes leveraging spectral features \cite{dwivedi2020generalization,kreuzer2021rethinking,lim2022sign} and graph features \cite{dwivedi2022LPE,chen2022SAT}.
Furthermore, standard global attention incurs quadratic computational costs $O(N^2)$ for a graph with $N$ nodes and $E$ edges, that limits GTs to small graphs with up to a few hundred nodes.

\begin{figure}[t]
  \centering
  \includegraphics[width=\textwidth]{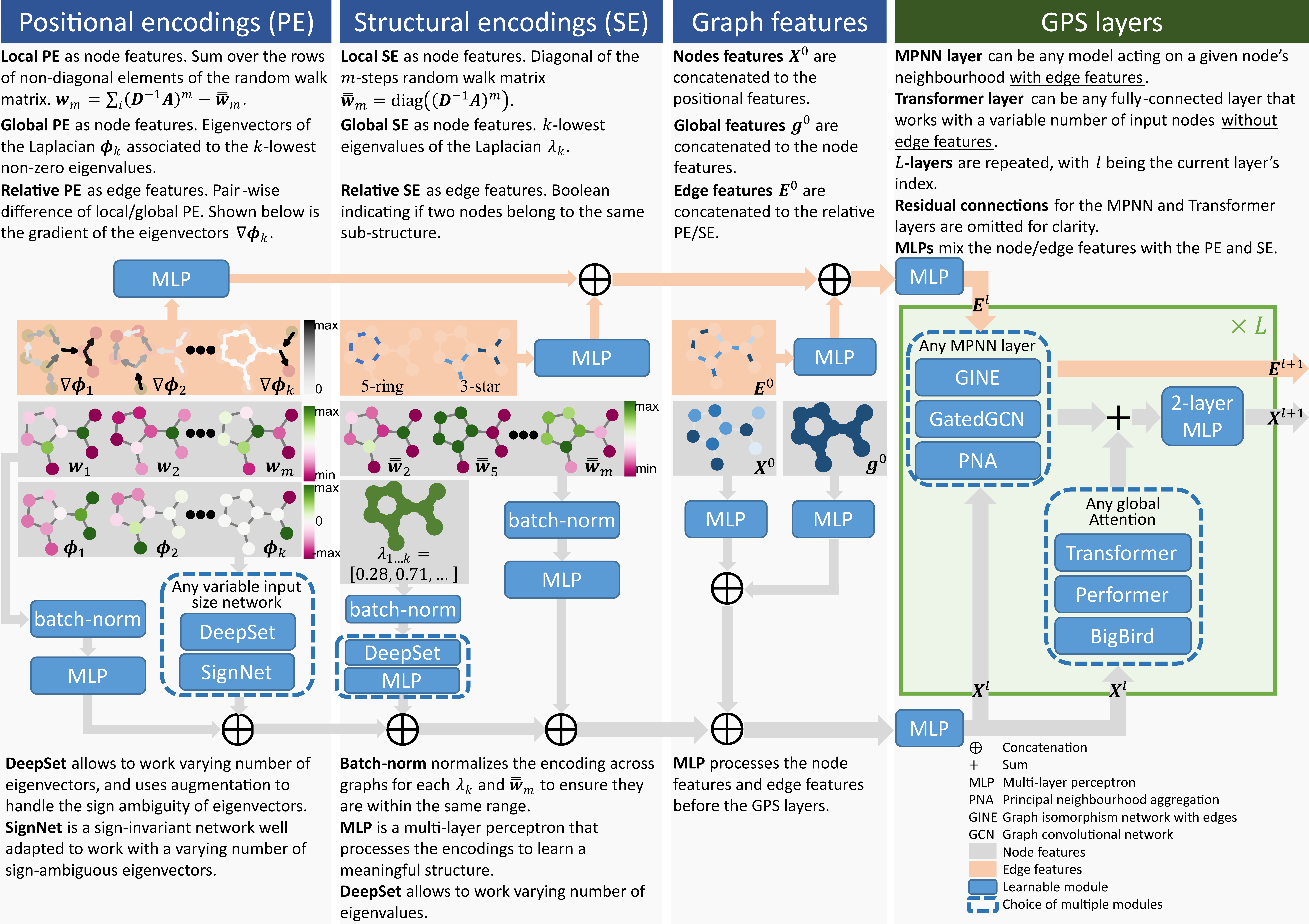}
  \caption{Modular \method graph Transformer,
  %Modular graph Transformer and \gtgym implementation
  with examples of PE and SE. Task specific layers for node/graph/edge-level predictions, such as pooling or output MLP, are omitted for simplicity.
  }
  \label{fig:visual-abstract}
\end{figure}

Whereas various GT models focus on particular node identifiability aspects, a principled framework for designing GTs is still missing. 
In this work, we address this gap and propose a recipe for building general, powerful, and scalable (GPS) graph Transformers.
The recipe defines (i) embedding modules responsible for aggregating \emph{positional encodings} (PE) and \emph{structural encodings} (SE) with the node, edge, and graph level input features;
% with input graph features on the node, edge, and graph levels
(ii) processing modules that employ a combination of local message passing and global attention layers (see Figure~\ref{fig:visual-abstract}).

The embedding modules organize multiple proposed PE and SE schemes into \emph{local} and \emph{global} levels serving as additional node features whereas positional and structural \emph{relative} features contribute to edge features. 
The processing modules define a computational graph that allows to balance between message-passing graph neural networks (MPNNs) and Transformer-like global attention, including attention mechanisms \emph{linear} in the number of nodes $O(N)$. 

To the best of our knowledge, application of efficient attention models has not yet been thoroughly studied in the graph domain, e.g., only one work~\cite{choromanski2021blocktoeplitz} explores the adaptation of Performer-style~\cite{DBLP:conf/iclr/ChoromanskiLDSG21} attention approximation on small graphs. Particular challenges emerge with explicit edge features that are incorporated as attention bias in fully-connected graph transformers~\cite{kreuzer2021rethinking, ying2021graphormer}.
Linear transformers do not materialize the attention matrix directly, hence incorporating edge features becomes a non-trivial task.
In this work, we hypothesize that explicit edge features are not necessary for the \emph{global graph attention} and adopt Performer~\cite{DBLP:conf/iclr/ChoromanskiLDSG21} and BigBird~\cite{DBLP:conf/nips/ZaheerGDAAOPRWY20} as exemplary linear attention mechanisms.

Our contributions are as follows. (i) Provide a general, powerful, scalable (\method) GT blueprint that incorporates positional and structural encodings with local message passing and global attention, visualized in Figure \ref{fig:visual-abstract}. (ii) Provide a better definition of PEs and SEs and organize them into \textit{local, global}, and \textit{relative} categories. (iii) Show that \method with linear global attention, e.g., provided by Performer \cite{DBLP:conf/iclr/ChoromanskiLDSG21} or BigBird \cite{DBLP:conf/nips/ZaheerGDAAOPRWY20}, scales to graphs with several thousand nodes and demonstrates competitive results even without explicit edge features within the attention module, whereas existing fully-connected GTs \cite{kreuzer2021rethinking, ying2021graphormer} are limited to graphs of up to few hundred nodes. (iv) Conduct an extensive ablation study that evaluates contribution of PEs, local MPNN, and global attention components in perspective of several benchmarking datasets. (v) Finally, following the success of GraphGym~\cite{you2020design} we implement the blueprint within a modular and performant \gtgym package.

%##############################################################################
% \vspace{-4pt}
\section{Related Work}
% \vspace{-4pt}
%##############################################################################

\myparagraph{Graph Transformers (GT).}
Considering the great successes of Transformers in natural language processing (NLP) \cite{vaswani2017attention, kalyan2021ammus_transformer_survey} and recently also in computer vision \cite{dascoli2021convit, han2022survey_vision_transformer, guo2021cmt}, it is natural to study their applicability in the graph domain as well. Particularly, they are expected to help alleviate the problems of over-smoothing and over-squashing \cite{alon2021on, topping2021understanding_ricci} in MPNNs, which are analogous to the vanishing gradients and lack of long-term dependencies in NLP. Fully-connected Graph Transformer~\cite{dwivedi2020generalization} was first introduced together with rudimentary utilisation of eigenvectors of the graph Laplacian as the node positional encoding (PE), to provide the otherwise graph-unaware Transformer a sense of nodes' location in the input graph.
Building on top of this work, SAN~\cite{kreuzer2021rethinking} implemented an invariant aggregation of Laplacian's eigenvectors for the PE, alongside conditional attention for real and virtual edges of a graph, which jointly yielded significant improvements.
% Building on top of this work, SAN implemented a learnable Laplacian PE alongside a separate head for the local/global aggregations, which yielded significant improvements \cite{kreuzer2021rethinking}.
Concurrently, Graphormer~\cite{ying2021graphormer,shi2022benchgraphormer} proposed using pair-wise graph distances (or 3D distances) to define relative positional encodings, with outstanding success on large molecular benchmarks.
Further, GraphiT~\cite{mialon2021graphit} used relative PE derived from diffusion kernels to modulate the attention between nodes. Finally, GraphTrans~\cite{jain2021graphtrans} proposed the first hybrid architecture, first using a stack of MPNN layers, before fully-connecting the graph.
Since, the field has continued to propose alternative GTs: SAT~\cite{chen2022SAT}, EGT~\cite{hussain2022EGT}, GRPE~\cite{park2022GRPE}.

\myparagraph{Positional and structural encodings.}
There have been many recent works on PE and SE, notably on Laplacian PE \cite{dwivedi2020generalization,kreuzer2021rethinking, beaini2021directional_dgn,wang2022equivstable,lim2022sign}, shortest-path-distance \cite{li2020distance,ying2021graphormer}, node degree centrality \cite{ying2021graphormer}, kernel distance \cite{mialon2021graphit}, random-walk SE \cite{dwivedi2022LPE}, structure-aware \cite{chen2022SAT, bouritsas2022improving_GSN,bodnar2021weisfeiler_CIN}, and more. Some works also propose dedicated networks to learn the PE/SE from an initial encoding \cite{kreuzer2021rethinking, dwivedi2022LPE, lim2022sign, chen2022SAT}. To better understand the different PE/SE and the contribution of each work, we categorize them in Table~\ref{tab:pe_se} and examine their effect in Section \ref{sec:need_pe_se}. In most cases, PE/SE are used as soft bias, meaning they are simply provided as input features. But in other cases, they can be used to direct the messages \cite{beaini2021directional_dgn} or create \textit{bridges} between distant nodes \cite{koutis2019spectral, topping2021understanding_ricci}.

\myparagraph{Linear Transformers.}
The \emph{quadratic} complexity of attention in the original Transformer architecture~\cite{vaswani2017attention} motivated the search for more efficient attention mechanisms that would scale \emph{linearly} with the sequence length. 
Most of such \emph{linear transformers} are developed for language modeling tasks, e.g., Linformer~\cite{Wang2020LinformerSW}, Reformer~\cite{DBLP:conf/iclr/KitaevKL20}, Longformer~\cite{DBLP:journals/corr/abs-2004-05150}, Performer~\cite{DBLP:conf/iclr/ChoromanskiLDSG21}, BigBird~\cite{DBLP:conf/nips/ZaheerGDAAOPRWY20}, and have a dedicated Long Range Arena benchmark~\cite{tay2021long} to study the limits of models against extremely long input sequences.
Pyraformer~\cite{liu2022pyraformer} is an example of a linear transformer for time series data, whereas S4~\cite{gu2022efficiently} is a more general signal processing approach that employs the state space model theory without the attention mechanism. 
In the graph domain, linear transformers are not well studied. \citet{choromanski2021blocktoeplitz} are the first to adapt Performer-style attention kernelization to small graphs.

%##############################################################################
\vspace{-3pt}
\section{Methods}
\vspace{-3pt}
%##############################################################################
In this work we provide a general, powerful, scalable (\method) architecture for graph Transformers, following our 3-part recipe presented in Figure~\ref{fig:visual-abstract}. We begin by categorization of existing positional (PE) and structural encodings (SE), a necessary ingredient for graph Transformers. Next, we analyse how these encodings also increase expressive power of MPNNs. The increased expressivity thus provides double benefit to our hybrid MPNN+Transformer architecture, which we introduce in Section~\ref{sec:gps_layer}. Last but not least, we provide an extensible implementation of \method in \gtgym package, built on top of PyG~\cite{FeyLenssen2019PyG} and GraphGym~\cite{you2020design}.
% The key ideas of our current work is to provide a modular/general, powerful/expressive, and fast/scalable architecture for graph Transformers. This is achieved from the architecture presented in Figure \ref{fig:visual-abstract}, with more details given below.

\vspace{-3pt}
\subsection{Modular positional and structural encodings}
\vspace{-3pt}
\label{sec:pe_se}
One of our contribution is to provide a modular framework for PE/SE. It was shown in previous works that they are one of the most important factors in driving the performance of graph Transformers. Thus, a better understanding and organization of the PE and SE will aid in building of a more modular architecture and in guiding of the future research.

We propose to organize the PE and SE into 3 categories: \textbf{local}, \textbf{global} and \textbf{relative} in order to facilitate the integration within the pipeline and facilitate new research directions. They are presented visually in Figure \ref{fig:visual-abstract}, with more details in Table \ref{tab:pe_se}. Although PE and SE can appear similar to some extent, they are different yet complementary. PE gives a notion of distance, while SE gives a notion of structural similarity. One can always infer certain notions of distance from large structures, or certain notions of structure from short distances, but this is not a trivial task, and the objective of providing PE and SE remains distinct, as discussed in the following subsections.

Despite presenting a variety of possible functions, we focus our empirical evaluations on the \textbf{global PE}, \textbf{relative PE} and \textbf{local SE} since they are known to yield significant improvements. We leave the empirical evaluation of other encodings for future work.

\myparagraph{Positional encodings (PE)}  are meant to provide an idea of the \textit{position in space} of a given node within the graph. Hence, when two nodes are close to each other within a graph or subgraph, their PE should also be close. A common approach is to compute the pair-wise distance between each pairs of nodes or their eigenvectors as proposed in \cite{li2020distance, ying2021graphormer, kreuzer2021rethinking, wang2022equivstable}, but this is not compatible with linear Transformers as it requires to materialize the full attention matrix~\cite{DBLP:conf/iclr/ChoromanskiLDSG21}. Instead, we want the PE to either be features of the nodes or real edges of the graph, thus a better fitting solution is to use the eigenvectors of the graph Laplacian or their gradient \cite{dwivedi2020benchmarking, beaini2021directional_dgn, kreuzer2021rethinking}. See Table~\ref{tab:pe_se} for more PE examples.

\begin{table}[t]
\vspace*{-4pt}
\fontsize{8.5pt}{8.5pt}\selectfont
\caption{The proposed categorization of positional encodings (\PE) and structural encodings (\SE). Some encodings are assigned to multiple categories in order to show their multiple expected roles.}
\vspace*{3pt}
\label{tab:pe_se}
\begin{adjustwidth}{-2.5 cm}{-2.5 cm}\centering
\setlength\tabcolsep{2pt} % default value: 6pt
\begin{tabular}{@{}m{0.13\textwidth}m{0.36\textwidth}m{0.53\textwidth}@{}}\toprule
\multicolumn{1}{c}{Encoding type} & \multicolumn{1}{c}{Description} & \multicolumn{1}{c}{Examples}\\
\midrule
\textbf{Local \PE} \newline \textit{node features}    &
Allow a node to know its position and role within a local cluster of nodes. 
\newline
\textit{Within a \textbf{cluster}, the closer two nodes are to each other, the closer their local PE will be, such as the position of a word in a sentence (not in the text).}
&
    % EXAMPLES LOCAL PE
    \begin{itemize}[leftmargin=*,topsep=2pt,itemsep=0pt,parsep=0pt,partopsep=2pt]
    \item Sum each column of non-diagonal elements of the $m$-steps random walk matrix.
    \item Distance between a node and the centroid of a cluster containing the node.
    \vspace*{-\baselineskip}
    \end{itemize}
\\
\midrule
\textbf{Global \PE} \newline \textit{node features}   & 
Allow a node to know its global position within the graph.
\newline
\textit{Within a \textbf{graph}, the closer two nodes are, the closer their global PE will be, such as the position of a word in a text.}
& 
    % EXAMPLES GLOBAL PE
    \begin{itemize}[leftmargin=*,topsep=2pt,itemsep=0pt,parsep=0pt,partopsep=2pt]
    \item Eigenvectors of the Adjacency, Laplacian \cite{dwivedi2020benchmarking, kreuzer2021rethinking} or distance matrices.
    \item SignNet~\cite{lim2022sign} (includes aspects of relative PE and local SE).
    \item Distance from the graph's centroid.
    \item Unique identifier for each connected component of the graph.
    \vspace*{-\baselineskip}
    \end{itemize}
\\
\midrule
\textbf{Relative \PE} \newline \textit{edge features} & 
Allow two nodes to understand their distances or directional relationships.
\newline
\textit{Edge embedding that is correlated to the distance given by any global or local PE, such as the distance between two words.}
& 
    % EXAMPLES RELATIVE PE
    \begin{itemize}[leftmargin=*,topsep=2pt,itemsep=0pt,parsep=0pt,partopsep=2pt]
    % \item Pair-wise node distances from heat kernels, random-walks, Green's function, graph geodesic \cite{beaini2021directional_dgn, kreuzer2021rethinking, mialon2021graphit}, or any \mbox{local/global} PE.
    \item Pair-wise node distances~\cite{li2020distance, beaini2021directional_dgn, kreuzer2021rethinking, ying2021graphormer, mialon2021graphit} based on shortest-paths, heat kernels, random-walks, Green's function, graph geodesic, or any \mbox{local/global} PE.
    \item Gradient of eigenvectors \cite{beaini2021directional_dgn, kreuzer2021rethinking} or any local/global PE.
    \item PEG layer~\cite{wang2022equivstable} with specific node-wise distances. % (includes  aspects of global PE).
    \item Boolean indicating if two nodes are in the same cluster.
    \vspace*{-\baselineskip}
    \end{itemize}
    \\
% \midrule[1pt]
\specialrule{.1em}{.05em}{.05em}
\textbf{Local \SE} \newline \textit{node features}    &
Allow a node to understand what sub-structures it is a part of. 
\newline
\textit{Given an SE of radius $m$, the more similar the $m$-hop subgraphs around two nodes are, the closer their local SE will be.}
&
    % EXAMPLES LOCAL SE
    \begin{itemize}[leftmargin=*,topsep=2pt,itemsep=0pt,parsep=0pt,partopsep=2pt]
    \item Degree of a node \cite{ying2021graphormer}.
    \item Diagonal of the $m$-steps random-walk matrix \cite{dwivedi2022LPE}.
    \item Time-derivative of the heat-kernel diagonal (gives the degree at $t=0$).
    \item Enumerate or count predefined structures such as triangles, rings, etc. \cite{bouritsas2022improving_GSN, zhao2021stars}.
    \item Ricci curvature \cite{topping2021understanding_ricci}.
    \vspace*{-\baselineskip}
    \end{itemize}
\\
\midrule
\textbf{Global \SE} \newline \textit{graph features}   & 
Provide the network with information about the global structure of the graph. 
\newline
\textit{The more similar two graphs are, the closer their global SE will be.}
& 
    % EXAMPLES GLOBAL SE
    \begin{itemize}[leftmargin=*,topsep=2pt,itemsep=0pt,parsep=0pt,partopsep=2pt]
    \item Eigenvalues of the Adjacency or Laplacian matrices \cite{kreuzer2021rethinking}.
    \item Graph properties: diameter, girth, number of connected components, \# of nodes, \# of edges, nodes-to-edges ratio.
    \vspace*{-\baselineskip}
    \end{itemize}
\\
\midrule
\textbf{Relative \SE} \newline \textit{edge features} & 
Allow two nodes to understand how much their structures differ.
\newline
\textit{Edge embedding that is correlated to the difference between any local SE.}
& 
    % EXAMPLES RELATIVE SE
    \begin{itemize}[leftmargin=*,topsep=2pt,itemsep=0pt,parsep=0pt,partopsep=2pt]
    % \item Distance Encoding~\cite{li2020distance} using shortest-path-distances~\cite{ying2021graphormer} or landing probabilities.
    % \item PEG-layer design~\cite{wang2022equivstable} with Deepwalk~\cite{perozzi2014deepwalk}.
    \item Pair-wise distance, encoding, or gradient of any local SE.
    \item Boolean indicating if two nodes are in the same sub-structure \cite{bodnar2021weisfeiler_CIN} (similar to the gradient of sub-structure enumeration).
    \vspace*{-\baselineskip}
    \end{itemize}
    \\
\bottomrule
\end{tabular}
\end{adjustwidth}
\vspace*{-10pt}
\end{table}

\myparagraph{Structural encodings (SE)} are meant to provide an embedding of the structure of graphs or subgraphs to help increase the expressivity and the generalizability of graph neural networks (GNN). Hence, when two nodes share similar subgraphs, or when two graphs are similar, their SE should also be close. Simple approaches are to identify pre-defined patterns in the graphs as one-hot encodings, but they require expert knowledge of graphs \cite{bouritsas2022improving_GSN, bodnar2021weisfeiler_CIN}. Instead, using the diagonal of the $m$-steps random-walk matrix encodes richer information into each node \cite{dwivedi2022LPE}, such as for odd $m$ it can indicate if a node is a part of an $m$-long cycle. Structural encodings can also be used to define the \textit{global} graph structure, for instance using the eigenvalues of the Laplacian, or as \textit{relative} edge features to identify if nodes are contained within the same clusters, with more examples in Table~\ref{tab:pe_se}.

\subsection{Why do we need PE and SE in MPNN?}
\label{sec:need_pe_se}
As reviewed earlier, several recent GNNs make use of positional encodings (PE) and structural encodings (SE) as soft biases to improve the model expressivity (summarized in Table \ref{tab:pe_se}), which also leads to better generalization. In this section, we present an examination of PE and SE by showing how message-passing networks, despite operating on the graph structure, remain blind to the information encapsulated by the PE and SE.

\begin{wrapfigure}{r}{0.28\textwidth}
    \vspace{-6pt}
    \begin{subfigure}[t]{\linewidth}
    \centering
    \includegraphics[width=1\linewidth]{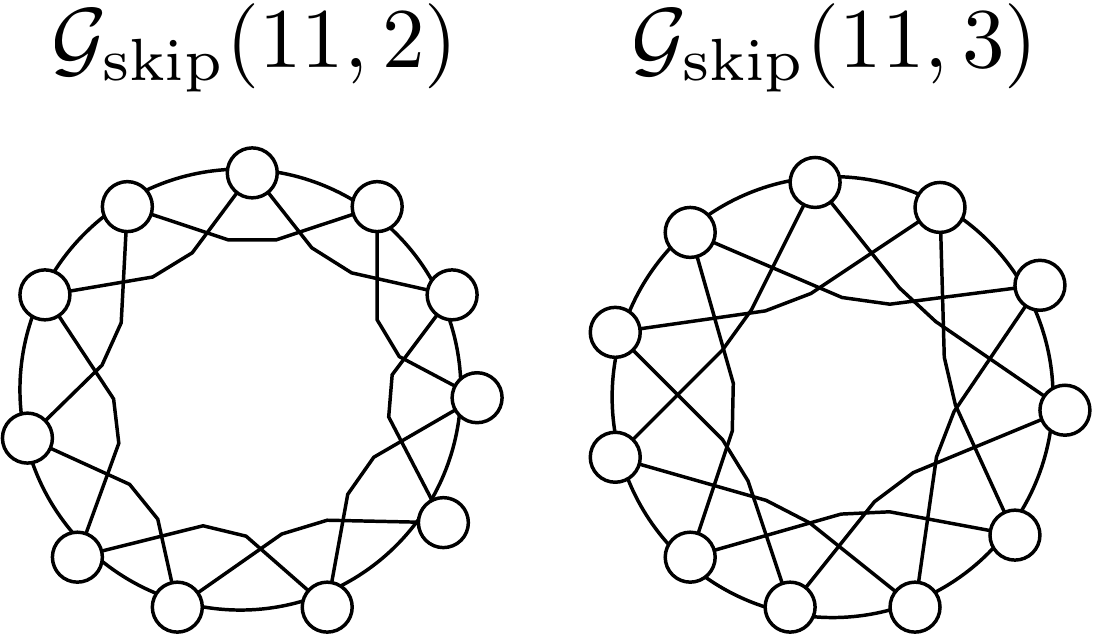}
    \subcaption{Circular Skip Link graphs}
    \label{fig:pe_se_main:a}
    \end{subfigure}
    \begin{subfigure}[t]{\linewidth}
    \centering
    \vspace{5pt}
    \includegraphics[width=0.8\linewidth]{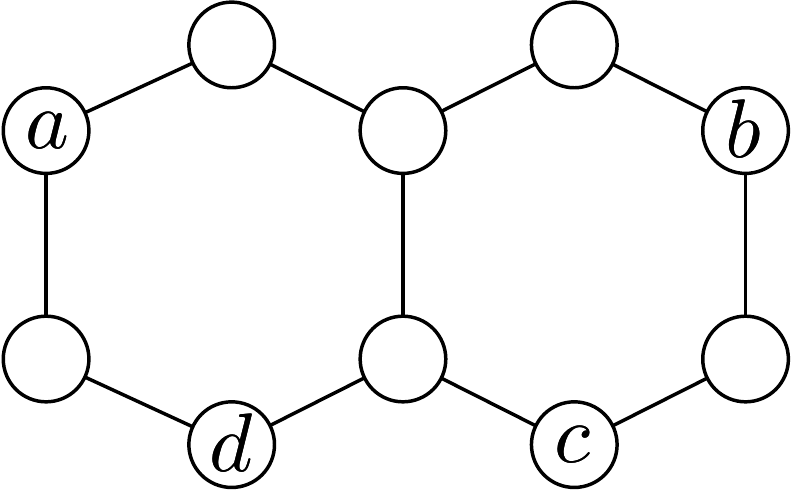}
    \vspace{-2pt}
    \subcaption{Decalin molecular graph}
    \label{fig:pe_se_main:b}
    \end{subfigure}
    % \caption{Example graphs with anonymous nodes: (a) A pair of Circular Skip Link graphs where the nodes have skip links of 2 and 3 respectively. (b) A Decalin molecular graph that is composed of two rings of all Carbon atoms, thus has no distinguishing node features.
    % }
    \caption{Example graphs with anonymous nodes without distinguishing features.
    }\label{fig:pe_se_main}
    \vspace*{-15pt}
\end{wrapfigure}

\myparagraph{1-Weisfeiler-Leman test (1-WL).}
It is well known that standard MPNNs are as expressive as the 1-WL test, meaning that they fail to distinguish non-isomorphic graphs under a 1-hop aggregation. We argue that the selected \textit{local}, \textit{global} and \textit{relative} PE/SE allow MPNNs to become more expressive than the 1-WL test, thus making them fundamentally more expressive at distinguishing between nodes and graphs.
To this end, we study the following two types of graphs (Figure~\ref{fig:pe_se_main} and Appendix~\ref{app:WL}).
%To this end, we study two types of graphs shown in Figures~\ref{fig:pe_se_main}~and~\ref{fig:pe_se_example}, with further details provided in Appendix~\ref{app:WL}.

% \begin{figure}[t]
%     \vspace{-15pt}
%   \centering
%   \includegraphics[width=0.7\textwidth]{pe_se_example.pdf}
%   \vspace{-3pt}
%   \caption{Example graphs with anonymous nodes:
%   %\textit{i.e.}, nodes do not have any distinguishing node features:
%   (a) A pair of Circular Skip Link (CSL) graphs \cite{murphy2019relational} where the nodes have skip links of 2 and 3 respectively. (b) A Decalin molecular graph that is composed of two rings of all Carbon atoms, thus has no distinguishing node features.
%   }
%   \label{fig:pe_se_example_small}
% \end{figure}

\myparagraph{Circular Skip Link (CSL) graph.}
In a CSL graph-pair \cite{murphy2019relational}, 
% Figures~\ref{fig:pe_se_main:a}~and~\ref{fig:pe_se_example:a}, 
we want to be able to distinguish the two non-isomorphic graphs. Since the 1-WL algorithm produces the same color for every node in both graphs, also every MPNN will fail to distinguish them. However, using a \textit{global} PE (e.g., Laplacian PE \cite{dwivedi2020benchmarking}) assigns each node a unique initial color and makes the CSL graph-pair distinguishable. This demonstrates that an MPNN cannot learn such a PE from the graph structure alone. Next, using a \textit{local} SE (e.g., diagonals of $m$-steps random walk) can successfully capture the difference in the skip links of the two graphs \cite{loukas2020graph}, resulting in their different node coloring~\cite{dwivedi2022LPE}.

\myparagraph{Decalin molecule.}
In the bicyclic Decalin graph, Figures~\ref{fig:pe_se_main:b}~and~\ref{fig:pe_se_example:b}, the node $a$ is isomorphic to node $b$, and so is the node $c$ to node $d$. A \mbox{1-WL} coloring of the nodes, and analogously MPNN, would generate one color for the nodes $a,b$ and another color for $c,d$. The same applies to the aforementioned \textit{local} SE \cite{dwivedi2022LPE}. In case of link prediction, this causes potential links $(a,d)$ and $(b,d)$ to be indistinguishable \cite{zhang2021labeling}. 
%However, certain tasks could require distinguishing between a potential link $(a,d)$ with another potential link $(b,d)$ with $(b,d)$ \cite{zhang2021labeling}. 
Using a distance-based \textit{relative} PE on the edges or an eigenvector-based \textit{global} PE, however, would allow to differentiate the two links.
% It henceforth justifies the theoretical need to encode PE and SE at any or all of the 3 levels discussed to make the MPNN strictly powerful than 1-WL.

% More details on the 1-WL test and how PE and SE help distinguishing the CSL graph and the nodes of the Decalin molecule are presented in Appendix~\ref{app:WL}.

\subsection{GPS layer: an MPNN+Transformer hybrid}\label{sec:gps_layer}
% In this section, we argue the different motivations behind using a hybrid MPNN+Transformer layer, which goes against most of the published methods. First, such layer facilitates scaling of the GT linearly to the number of nodes and edges, secondly it helps avoiding over-smoothing and over-squashing in the first layers. Finally, we present the equations for the update function.
% \vd{(VD: This paragraph needs re-writing as the scalability part is moved up. An alternative is drafted as follows.)}

In this section we introduce the GPS layer, which is a hybrid MPNN+Transformer layer. First we argue how it alleviates the limitation of a closely related work. Next, we list the layer update equations which can be instantiated with a variety of MPNN and Transformer layers. Finally, we present its characteristics in terms of modularity, scalability and expressivity.

% Most of the proposed methods in the literature, such as GT~\cite{dwivedi2020generalization}, SAN~\cite{kreuzer2021rethinking}, Graphormer~\cite{ying2021graphormer} use a model based purely on full self-attention, while only GraphTrans~\cite{jain2021graphtrans} uses a hybrid architecture. Yet, even GraphTrans first starts with MPNN module before applying the full self-attention stack.

% \myparagraph{Scalability and expressivity.} As mentioned in previous sections, processing the edges separately in the MPNN allows for the use of linear Transformers. Indeed, the MPNN complexity is $O(E)$ and the linear Transformers are $O(N)$. For sparse graphs such as molecular graphs, regular graphs, and knowledge graphs, the edges are practically proportional to the nodes $E = \Theta(N)$, meaning the entire complexity becomes linear to the number of nodes $O(N)$.

\myparagraph{Preventing early smoothing.} Why not use an architecture like GraphTrans \cite{jain2021graphtrans} comprising of a few layers of MPNNs before the Transformer? Since MPNNs are limited by problems of over-smoothing, over-squashing, and low expressivity against the WL test \cite{alon2021on,topping2021understanding_ricci}, these layers could \textit{irreparably} fail to keep some information in the early stage. Although they could make use of PE/SE or more expressive MPNNs \cite{beaini2021directional_dgn, dwivedi2022LPE}, they are still likely to lose information. An analogous 2-stage strategy was successful in computer vision \cite{dascoli2021convit, guo2021cmt} thanks to the high expressivity of convolutional layers on grids, but we do not expect it to achieve the same success on graphs due to the limitations of message-passing.

\newcommand{\mpnn}{\texttt{MPNN}\xspace}
\newcommand{\attn}{\texttt{GlobalAttn}\xspace}
\newcommand{\X}{\mathbf{X}}
\newcommand{\E}{\mathbf{E}}
\newcommand{\A}{\mathbf{A}}
\myparagraph{Update function.}
At each layer, the features are updated by aggregating the output of an MPNN layer with that of a global attention layer, as shown in Figures~\ref{fig:visual-abstract}~and~\ref{fig:gps_layer}, and described by the equations below. Note that the edge features are only passed to the \mpnn layer, and that residual connections with batch normalization~\cite{ioffe2015batchnorm} are omitted for clarity. Both the \mpnn and \attn layers are modular, i.e., \mpnn can be any function that acts on a local neighborhood and \attn can be any fully-connected layer.
\vspace{-3pt}
\begin{eqnarray}
    \X^{\ell+1}, \E^{\ell+1} &=& \texttt{GPS}^{\ell} \left( \X^{\ell}, \E^{\ell}, \A \right)\\
    \textrm{computed as} \ \ \ \ 
    \X^{\ell+1}_M, \ \E^{\ell+1} &=& \mpnn_e^{\ell} \left(\X^{\ell}, \E^{\ell}, \A \right),\\
    \X^{\ell+1}_T
    &=&  \attn^{\ell} \left(\X^{\ell}  \right),\\
    \X^{\ell+1} &=&
    \texttt{MLP}^{\ell}\left(\X^{\ell+1}_M + \X^{\ell+1}_T\right),
    \label{eqn:layer_equation}
\end{eqnarray}
where $\A \in \mathbb{R}^{N \times N}$ is the adjacency matrix of a graph with $N$ nodes and $E$ edges; $\X^{\ell} \in \mathbb{R}^{N \times d_\ell}, \E^{\ell} \in \mathbb{R}^{E \times d_\ell}$ are the $d_\ell$-dimensional node and edge features, respectively; $\mpnn_e^{\ell}$ and $\attn^{\ell}$ are instances of an MPNN with edge features and of a global attention mechanism at the $\ell$-th layer with their corresponding learnable parameters, respectively; $\texttt{MLP}^{\ell}$ is a 2-layer MLP block.

\myparagraph{Modularity} is achieved by allowing drop-in replacement for a number of module choices, including the initial PE/SE types, the networks that processes those PE/SE, the MPNN and global attention layers that constitute a GPS layer, and the final task-specific prediction head. Further, as research advances in different directions, \gtgym allows to easily implement new PE/SE and other layers.

\myparagraph{Scalability} is achieved by allowing for a computational complexity linear in both the number of nodes and edges $O(N+E)$; excluding the potential precomputation step required for various PE, such as Laplacian eigen-decomposition.
By restricting the PE/SE to real nodes and edges, and by excluding the edge features from the global attention layer, we can avoid materializing the full quadratic attention matrix. Therefore we can utilize a linear Transformer with $O(N)$ complexity, while the complexity of an MPNN is $O(E)$. For sparse graphs such as molecular graphs, regular graphs, and knowledge graphs, the edges are practically proportional to the nodes $E = \Theta(N)$, meaning the entire complexity can be considered linear in the number of nodes $O(N)$.
Empirically, even on small molecular graphs, our architecture reduces computation time compared to other GT models, e.g., a model of \textasciitilde6M parameters requires 196s per epoch on the ogbg-molpcba~\cite{hu2020ogb} dataset, compared to 883s for SAN \cite{kreuzer2021rethinking} on the same GPU type.

\myparagraph{Expressivity} in terms of sub-structure identification and the Weisfeiler-Leman (WL) test is achieved via providing a rich set of PE/SE, as proposed in various works \cite{beaini2021directional_dgn, kreuzer2021rethinking, dwivedi2022LPE, bodnar2021weisfeiler_CIN, bouritsas2022improving_GSN} and detailed in Section \ref{sec:pe_se}. Further, the Transformer allows to resolve the expressivity bottlenecks caused by over-smoothing \cite{kreuzer2021rethinking} and over-squashing \cite{alon2021on} by allowing information to spread across the graph via full-connectivity. Finally, in Section~\ref{sec:theory}, we demonstrate that, given the right components, the proposed architecture does not lose edge information and is a universal function approximator on graphs.

% \vspace{-3pt}
\subsection{Theoretical expressivity}\label{sec:theory}
% \vspace{-3pt}
%In this section, we argue that the proposed architecture is a universal approximator for graphs.
%First, as shown in previous works, the model is more expressive than the 1-WL test since it uses eigenvectors PE \cite{beaini2021directional_dgn,kreuzer2021rethinking} and random-walk SE \cite{dwivedi2022LPE}. 
In this section, we first discuss how the MPNN layer allows to propagate edge and neighbor information on the nodes. Then, we show that the proposed model is a universal function approximator on graphs, similarly to the SAN architecture \cite{kreuzer2021rethinking}.

% \subsubsection{Preserving edge information in the Transformer layer}
% \label{sec:edge_info}
\myparagraph{Preserving edge information in the Transformer layer.}
Most GTs do not fully utilize edge features of the input graph. The Graph Transformer~\cite{dwivedi2020generalization}, SAN~\cite{kreuzer2021rethinking} and Graphormer~\cite{ying2021graphormer} only use edge features to condition the attention between a pair of nodes, that is, they influence the attention gating mechanism but are not explicitly involved in updating of the node representations. GraphiT~\cite{mialon2021graphit} does not consider edge features at all. Recent 2-step methods GraphTrans~\cite{jain2021graphtrans} and SAT~\cite{chen2022SAT} can use edge features in their first MPNN step, however this step is applied only once and typically includes several $k$ rounds of message passing. Therefore this latter approach may suffer from initial over-smoothing, as $k$-hop neighborhoods together with the respective edge features need to be represented in a fixed-sized node representation.

On the other hand, in \method, interleaving one round of local neighborhood aggregation via an MPNN layer with global self-attention mechanism reduces the initial representation bottleneck and enables iterative local and global interactions. In the attention, the key-query-value mechanism only explicitly depends on the node features, but assuming efficient representation encoding by the MPNN, the node features can implicitly encode edge information, thus edges can play a role in either the key, query, or values. In Appendix \ref{app:edge-in-attention}, we give a more formal argument on how, following an MPNN layer, node features can encode edge features alongside information related to node-connectivity.

% \subsubsection{Universal function approximator on graphs}
% \label{sec:universal}
\myparagraph{Universal function approximator on graphs.}
\citet{kreuzer2021rethinking}[Sec.~3.5] demonstrated the universality of graph Transformers. It was shown that, given the full set of Laplacian eigenvectors, the model was a universal function approximator on graphs and could provide an approximate solution to the isomorphism problem, making it more powerful than any Weisfeiler-Leman (WL) isomorphism test given enough parameters.
Here, we argue that the same holds for our architecture since we can also use the full set of eigenvectors, and since all edge information can be propagated to the nodes.

% \textcolor{blue}{TODO -- mention how this is WL-inf, same as SAN}

% \lr{Argument / proof about WL expressivity of our model -- Reuse argument from SAN and Vijay's Learnable PE about RWSE.}
% \db{Don't think we need this, I already stated earlier the proofs from SAN and RWSE that are valid and we are >1-WL, without proof.}

%##############################################################################
% \vspace{-3pt}
\section{Experiments}
% \vspace{-3pt}
%##############################################################################
We perform ablation studies on 4 datasets to evaluate the contribution of the message-passing module, the global attention module, and the positional or structural encodings. Then, we evaluate \method on a diverse set of 11 benchmarking datasets, and show state-of-the-art (SOTA) results in many cases.

We test on datasets from different sources to ensure diversity, providing their detailed description in Appendix~\ref{app:datasets}. From the Benchmarking GNNs~\cite{dwivedi2020benchmarking}, we test on the ZINC, PATTERN, CLUSTER, MNIST, CIFAR10. From the open graph benchmark (OGB)~\cite{hu2020ogb}, we test on all graph-level datasets: ogbg-molhiv, ogbg-molpcba, ogbg-code2, and ogbg-ppa, and from their large-scale challenge we test on the OGB-LSC PCQM4Mv2~\cite{hu2021ogblsc}. Finally, we also select MalNet-Tiny~\cite{freitas2021malnet} with 5000 graphs, each of up to 5000 nodes, since the number of nodes provide a scaling challenge for Transformers.

\subsection{Ablation studies}\label{sec:ablations}
In this section, we evaluate multiple options for the three main components of our architecture in order to gauge their contribution to predictive performance and to better guide dataset-specific hyper-parameter optimization. First, we quantify benefits of the considered global-attention modules in 4 tasks. Then, we note that the MPNN layer is an essential part for high-performing models, and identify the layer type most likely to help. Finally, we observe when different global PE or local SE provide significant boost in the performance.
All ablation results are averaged over multiple random seeds and summarized in Table~\ref{fig:ablations}, with additional information available in Appendix~\ref{app:exp_ablations}.

\begin{table}[t]
\caption{Summary of the ablation studies. Details of the architectural choices, parameters, standard deviation, and computation times are presented in Appendix~\ref{app:exp_ablations}.
}
\label{fig:ablations}
\newcommand{\myfont}[1]{\fontsize{7pt}{7pt}\selectfont{#1}}
\newcolumntype{C}{>{\centering\arraybackslash}m{1.1cm}}
\centering
\begin{subfigure}[t]{.49\textwidth}
  \subcaption{Ablation of the Transformer and MPNN layers. We observe a major drop when using only a Transformer without an MPNN. Further, most datasets benefit from using a Transformer, without any negative impact.}
  \label{tab:ablation-mpnn-transformer}
  \centering
  \fontsize{8pt}{8pt}\selectfont
  \setlength\tabcolsep{1.4pt} % default value: 6pt
  \renewcommand{\arraystretch}{1.2}
  \begin{tabular}{@{}l l l C >{\centering\arraybackslash}m{1.15cm} C C @{}}\toprule
    \multicolumn{3}{c}{\multirow{3}{*}{Ablation}} &\multirow{2}{*}{ZINC} &\myfont{PCQM4Mv2} &\multirow{2}{*}{CIFAR10} &MalNet \\[-2pt]
    & & & &\myfont{subset} & &\myfont{-Tiny} \\[-3pt]\cmidrule{4-7}
    & & &MAE $\downarrow$ &MAE $\downarrow$ &Acc. $\uparrow$ &Acc. $\uparrow$ \\[-2pt]\midrule
    \multirow{4}{*}{\rotatebox[origin=c]{90}{\textbf{Global}}} &
    \multirow{4}{*}{\rotatebox[origin=c]{90}{\textbf{Attention}}} &
    \emph{none} &\cellcolor[HTML]{78c8a1}0.070 &\cellcolor[HTML]{8bceae}0.1213 &\cellcolor[HTML]{d8ebe1}69.95 &\cellcolor[HTML]{9cd5b9}92.23 \\
    & &Transformer &\cellcolor[HTML]{78c8a1}0.070 &\cellcolor[HTML]{7cc9a4}0.1159 &\cellcolor[HTML]{97d3b6}72.31 &\cellcolor[HTML]{78c8a1}93.50 \\
    & &Performer &\cellcolor[HTML]{7cc9a3}0.071 &\cellcolor[HTML]{78c8a1}0.1142 &\cellcolor[HTML]{c4e4d4}70.67 &\cellcolor[HTML]{90d1b2}92.64 \\
    & &BigBird &\cellcolor[HTML]{7cc9a3}0.071 &\cellcolor[HTML]{91d1b2}0.1237 &\cellcolor[HTML]{c9e6d8}70.48 &\cellcolor[HTML]{99d4b7}92.34 \\\midrule
    % \parbox[t]{2mm}{\multirow{4}{*}{\rotatebox[origin=c]{90}{\textbf{Change}}}} &
    \multicolumn{2}{c}{\multirow{4}{*}{\rotatebox[origin=c]{90}{\textbf{MPNN}}}} &
    \emph{none} &\cellcolor[HTML]{f5f5f5}0.217 &\cellcolor[HTML]{f5f5f5}0.3294 &\cellcolor[HTML]{f5f5f5}68.86 &\cellcolor[HTML]{f5f5f5}73.90 \\
    & &GINE &\cellcolor[HTML]{78c8a1}0.070 &\cellcolor[HTML]{9ed5bb}0.1284 &\cellcolor[HTML]{b8dfcc}71.11 &\cellcolor[HTML]{9bd5b8}92.27 \\
    & &GatedGCN &\cellcolor[HTML]{badfcd}0.086 &\cellcolor[HTML]{7cc9a4}0.1159 &\cellcolor[HTML]{97d3b6}72.31 &\cellcolor[HTML]{90d1b2}92.64 \\
    & &PNA &\cellcolor[HTML]{78c8a1}0.070 &\cellcolor[HTML]{c0e2d1}0.1409 &\cellcolor[HTML]{78c8a1}73.42 &\cellcolor[HTML]{abdbc4}91.67 \\
    \bottomrule
    \end{tabular}
\end{subfigure}%
\hfill%
\begin{subfigure}[t]{.49\textwidth}
  \subcaption{Ablation of the \PE and \SE types. RWSE provides consistent gains at relatively low computational cost, while SignNet\textsuperscript{DeepSets} is the single best performing encoding, albeit at increased computational cost.}
  \label{tab:ablation-pe-se}
  \centering
  \fontsize{8pt}{8pt}\selectfont
  \setlength\tabcolsep{1.4pt} % default value: 6pt
  \renewcommand{\arraystretch}{1.2}
    \begin{tabular}{@{}l l C >{\centering\arraybackslash}m{1.15cm} C C@{}}\toprule
    \multicolumn{2}{c}{\multirow{3}{*}{Ablation}} &\multirow{2}{*}{ZINC} &\myfont{PCQM4Mv2} &\multirow{2}{*}{CIFAR10} &MalNet \\[-2pt]
    & & &\myfont{subset} & &\myfont{-Tiny} \\[-3pt]\cmidrule{3-6}
    & &MAE $\downarrow$ &MAE $\downarrow$ &Acc. $\uparrow$ &Acc. $\uparrow$ \\[-2pt]\midrule
    \multirow{6}{*}{\rotatebox[origin=c]{90}{\textbf{PE / SE}}} &\emph{none} &\cellcolor[HTML]{d1e8dd}0.113 &\cellcolor[HTML]{f5f5f5}0.1355 &\cellcolor[HTML]{f5f5f5}71.49 &\cellcolor[HTML]{c0e2d1}92.64 \\
    &\textcolor{se_color}{RWSE} &\cellcolor[HTML]{78c8a1}0.070 &\cellcolor[HTML]{80cba6}0.1159 &\cellcolor[HTML]{b3ddc9}71.96 &\cellcolor[HTML]{addbc5}92.77 \\
    &\textcolor{pe_color}{LapPE} &\cellcolor[HTML]{d7eae1}0.116 &\cellcolor[HTML]{99d4b7}0.1201 &\cellcolor[HTML]{81cca8}72.31 &\cellcolor[HTML]{b1ddc8}92.74 \\
    &\textcolor{pe_color}{SignNet\textsuperscript{MLP}} &\cellcolor[HTML]{a1d6bc}0.090 &\cellcolor[HTML]{80caa6}0.1158 &\cellcolor[HTML]{d2e9dd}71.74 &\cellcolor[HTML]{cae6d8}92.57 \\
    &\textcolor{pe_color}{SignNet\textsuperscript{DeepSets}} &\cellcolor[HTML]{8acead}0.079 &\cellcolor[HTML]{78c8a1}0.1144 &\cellcolor[HTML]{78c8a1}72.37 &\cellcolor[HTML]{78c8a1}93.13 \\
    &\textcolor{pe_color}{PEG\textsuperscript{LapEig}} &\cellcolor[HTML]{f5f5f5}0.161 &\cellcolor[HTML]{9ed5ba}0.1209 &\cellcolor[HTML]{9fd6bb}72.10 &\cellcolor[HTML]{f5f5f5}92.27 \\
    \bottomrule
    \multicolumn{6}{l}{\fontsize{7.5pt}{7.5pt}\selectfont *Encodings are color-coded by their \textcolor{pe_color}{positional} or \textcolor{se_color}{structural} type.} \\
    \end{tabular}
\end{subfigure}
\end{table}

\paragraph{Global-Attention module.}
Here we consider global attention implemented as $O(N^2)$ key-query-value Transformer attention or linear-time attention mechanisms of Performer or BigBird. We notice in Table~\ref{tab:ablation-mpnn-transformer} that using a Transformer is always beneficial, except for the ZINC dataset where no changes are observed. This motivates our architecture and the hypothesis that long-range dependencies are generally important. We further observe that Performer falls behind Transformer in terms of the predictive performance, although it provides a gain over the baseline and the ability to scale to very large graphs. Finally, BigBird in our setting offers no significant gain, while also being slower than Performer (see Appendix~\ref{app:exp_ablations}).

Having no gain on the ZINC dataset is expected since the task is a combination of the computed octanol-water partition coefficient (cLogP) \cite{wildman1999prediction_logp} and the synthetic accessibility score (SA-score) \cite{ertl2009estimation_sascore}, both of which only count occurrences of local sub-structures. Hence, there is no need for a global connectivity, but a strong need for structural encodings.

\paragraph{Message-passing module.} Next, we evaluate the effect of various message-passing architectures, Table~\ref{tab:ablation-mpnn-transformer}. It is apparent that they are fundamental to the success of our method: removing the layer leads to a significant drop in performance across all datasets. Indeed, without an MPNN, the edge features are not taken into consideration at all. Additionally, without reinforcing of the local graph structure, the network can overfit to the PE/SE. This reiterates findings of \citet{kreuzer2021rethinking}, where considerably larger weights were assigned to the local attention.

We also find that although a vanilla PNA~\cite{corso2020principal_pna} generally outperforms GINE~\cite{hu2019strategies_GINE} and GatedGCN~\cite{bresson2017GatedGCN}, adding the PE and SE results in major performance boost especially for the GatedGCN. This is consistent with results of \citet{dwivedi2022LPE} and shows the importance of these encodings for gating.

Perhaps the necessity of a local message-passing module is due to the limited amount of graph data, and scaling to colossal datasets \cite{raffel2020t5} that we encounter in language and vision could change that. Indeed, the Graphormer architecture \cite{ying2021graphormer} was able to perform very well on the full PCQM4Mv2 dataset without a local module. However, even large Transformer-based language models \cite{chelombiev2021groupbert} and vision models \cite{han2022survey_vision_transformer} can benefit from an added local aggregation and outperform pure Transformers.

\paragraph{Positional/Structural Encodings.}
Finally, we evaluate the effects of various PE/SE schemes, Table~\ref{tab:ablation-pe-se}. We find them generally beneficial to downstream tasks, in concordance to the vast literature on the subject (see Table~\ref{tab:pe_se}). The benefits of the different encodings are very dataset dependant, with the random-walk structural encoding (RWSE) being more beneficial for molecular data and the Laplacian eigenvectors encodings (LapPE) being more beneficial in image superpixels. However, using SignNet with DeepSets encoding~\cite{lim2022sign} as an improved way of processing the LapPE seems to be consistently successful across tasks. We hypothesize that SignNet can learn structural representation using the eigenvectors, for example, to generate local heat-kernels that approximate random walks~\cite{andres2016heat}. Last but not least we evaluate PEG-layer design \cite{wang2022equivstable} with Laplacian eigenmap. %, applied to respective evaluated MPNN layer.

% We have observed in some experiments that using different PE/SE simultaneously can often be beneficial, which motivates our architectural choice. \db{For instance, we use it on the ogbg-molPCBA dataset to achieve SOTA results.} However, we leave the study of such combination for future work.

% \vspace{-3pt}
\subsection{Benchmarking \method}\label{sec:benchmarking}
% \vspace{-3pt}
% Reference results from existing papers (SAN, Graphormer, GraphiT, LSPE, SAT, EGT, GRPE) on available subset of the 11 datasets compared to best GPS configs.
We compare \method against a set of popular message-passing neural networks (GCN, GIN, GatedGCN, PNA, etc.), graph transformers (SAN, Graphormer, etc.), and other recent graph neural networks with SOTA results (CIN, CRaWL, GIN-AK+, ExpC). To ensure diverse benchmarking tasks, we use datasets from Benchmarking-GNNs \cite{dwivedi2020benchmarking}, OGB \cite{hu2020ogb} and its large-scale challenge \cite{hu2021ogblsc}, and Long-Range Graph Benchmark \cite{dwivedi2022LRGB}, with more details given in Appendix \ref{app:datasets}. We report the mean and standard deviation over 10 random seeds if not explicitly stated otherwise.

\begin{table}[t]
    \caption{Test performance in five benchmarks from \cite{dwivedi2020benchmarking}. Shown is the mean~±~s.d.~of 10 runs with different random seeds. Highlighted are the top \first{first}, \second{second}, and \third{third} results.
    %The metrics of compared methods are reprinted from their respective original publications or \citet{dwivedi2020benchmarking}.
    }
    \label{tab:results_benchgnns}
    \centering
    % \small
    \fontsize{8.5pt}{8.5pt}\selectfont
    \setlength\tabcolsep{4pt} % default value: 6pt
    \begin{tabular}{lcccccc}\toprule
    \multirow{2}{*}{\textbf{Model}} &\textbf{ZINC} &\textbf{MNIST} &\textbf{CIFAR10} &\textbf{PATTERN} &\textbf{CLUSTER} \\\cmidrule{2-6}
    &\textbf{MAE $\downarrow$} &\textbf{Accuracy $\uparrow$} &\textbf{Accuracy $\uparrow$} &\textbf{Accuracy $\uparrow$} &\textbf{Accuracy $\uparrow$} \\\midrule
    GCN \cite{kipf2016semi} &0.367 ± 0.011 &90.705 ± 0.218 &55.710 ± 0.381 &71.892 ± 0.334 &68.498 ± 0.976 \\
    GIN \cite{xu2018how} &0.526 ± 0.051 &96.485 ± 0.252 &55.255 ± 1.527 &85.387 ± 0.136 &64.716 ± 1.553 \\
    GAT \cite{velickovic2018GAT} &0.384 ± 0.007 &95.535 ± 0.205 &64.223 ± 0.455 &78.271 ± 0.186 &70.587 ± 0.447 \\
    GatedGCN \cite{bresson2017GatedGCN,dwivedi2020benchmarking} &0.282 ± 0.015 &97.340 ± 0.143 &67.312 ± 0.311 &85.568 ± 0.088 &73.840 ± 0.326 \\
    % GatedGCN-PE &0.214 ± 0.006 &-- &-- &86.508 ± 0.085 &76.082 ± 0.196 \\
    GatedGCN-LSPE \cite{dwivedi2022LPE} &0.090 ± 0.001 &-- &-- &-- &-- \\
    PNA \cite{corso2020principal_pna} &0.188 ± 0.004 &97.94 ± 0.12 &70.35 ± 0.63 &-- &-- \\
    DGN \cite{beaini2021directional_dgn} &0.168 ± 0.003 &-- &\first{72.838 ± 0.417} &86.680 ± 0.034 &-- \\
    GSN \cite{bouritsas2022improving_GSN} &0.101 ± 0.010 &-- &-- &-- &-- \\\midrule
    CIN \cite{bodnar2021weisfeiler_CIN} &\second{0.079 ± 0.006} &-- &-- &-- &-- \\
    CRaWl~\cite{toenshoff2021CRaWl} &0.085 ± 0.004 &\third{97.944 ± 0.050} &69.013 ± 0.259 &-- &-- \\
    GIN-AK+~\cite{zhao2021stars} &\third{0.080 ± 0.001} &-- &\third{72.19 ± 0.13} &\first{86.850 ± 0.057} &-- \\\midrule
    SAN \cite{kreuzer2021rethinking} &0.139 ± 0.006 &-- &-- &86.581 ± 0.037 &76.691 ± 0.65 \\
    Graphormer \cite{ying2021graphormer} &0.122 ± 0.006 &-- &-- &-- &-- \\
    K-Subgraph SAT \cite{chen2022SAT} &0.094 ± 0.008 &-- &-- &\second{86.848 ± 0.037} &\third{77.856 ± 0.104} \\
    EGT \cite{hussain2022EGT} &0.108 ± 0.009 &\first{98.173 ± 0.087} &68.702 ± 0.409 &\third{86.821 ± 0.020} &\first{79.232 ± 0.348} \\\midrule
    \method (ours) &\first{0.070 ± 0.004} & \second{98.051 ± 0.126} & \second{72.298 ± 0.356} &86.685 ± 0.059 & \second{78.016 ± 0.180} \\
    \bottomrule
    \end{tabular}
    % \vspace{-5pt}
\end{table}

\myparagraph{Benchmarking GNNs \cite{dwivedi2020benchmarking}.}
We first benchmark our method on 5 tasks from Benchmarking GNNs \cite{dwivedi2020benchmarking}, namely ZINC, MNIST, CIFAR10, PATTERN, and CLUSTER, shown in Table~\ref{tab:results_benchgnns}. We observe that our \method gives SOTA results on ZINC and the second best in 3 more datasets, showcasing the ability to perform very well on a variety of synthetic tasks designed to test the model expressivity.

\myparagraph{Open Graph Benchmark \cite{hu2020ogb}.}
Next, we benchmark on all 4 graph-level tasks from OGB, namely molhiv, molpcba, ppa, and code2, Table~\ref{tab:results_ogb}. On the molhiv dataset, we observed our model to suffer from overfitting, but to still outperform SAN, while other graph Transformers do not report results. For the molpcba, ppa, and code2, \method always ranks among the top 3 models, highlighting again the versatility and expressiveness of the \method approach. Further, \method outperforms every other GT on all 4 benchmarks, except SAT on code2.

\myparagraph{OGB-LSC PCQM4Mv2 \cite{hu2021ogblsc}.}
The large-scale PCQM4Mv2 dataset has been a popular benchmark for recent GTs, particularly due to Graphormer~\cite{ying2021graphormer} winning the initial challenge. We report the results in Table~\ref{tab:results_pcqm4m}, observing significant improvements over message-passing networks at comparable parameter budget. \method also outperforms GRPE~\cite{park2022GRPE}, EGT~\cite{hussain2022EGT}, and Graphormer \cite{ying2021graphormer} with less than half their parameters, and with significantly less overfitting on the training set.
Contrarily to Graphormer, we do not need to precompute spatial distances from approximate 3D molecular conformers \cite{ying2021first}, the RWSEs we utilize are graph-based only.

\myparagraph{MalNet-Tiny.}
The MalNet-Tiny~\cite{freitas2021malnet} dataset consists of function call graphs with up to 5,000 nodes. These graphs are considerably larger than previously considered inductive graph-learning benchmarks, which enables us to showcase scalability of \method to much larger graphs than prior methods. Our \method reaches $92.72\% \pm 0.7\mathrm{pp}$ test accuracy when using Performer global attention. Interestingly, using Transformer global attention leads to further improved \method performance, $93.36\% \pm 0.6\mathrm{pp}$ (based on 10 runs), albeit at the cost of doubled run-time. In both cases, we used comparable architecture to \citet{freitas2021malnet}, with 5 layers and 64 dimensional hidden node representation, and outperform their best GIN model with $90\%$ accuracy. See Appendix~\ref{app:exp_ablations} for \method ablation study on MalNet-Tiny.

\begin{table}[t]
    \caption{Test performance in graph-level OGB benchmarks~\cite{hu2020ogb}. Shown is the mean~±~s.d.~of 10 runs. Models that were first pre-trained on another dataset or use an ensemble are not included here.}
    \label{tab:results_ogb}
    \centering
    % \small
    \fontsize{8.5pt}{8.5pt}\selectfont
    \begin{tabular}{lccccc}\toprule
    \multirow{2}{*}{\textbf{Model}} &\textbf{ogbg-molhiv} &\textbf{ogbg-molpcba} &\textbf{ogbg-ppa} &\textbf{ogbg-code2} \\\cmidrule{2-5}
    &\textbf{AUROC $\uparrow$} &\textbf{Avg.~Precision $\uparrow$} &\textbf{Accuracy $\uparrow$} &\textbf{F1 score $\uparrow$} \\\midrule
    % GCN &0.7606 ± 0.0097 &0.2020 ± 0.0024 &0.6839 ± 0.0084 &0.1507 ± 0.0018 \\
    GCN+virtual node &0.7599 ± 0.0119 &0.2424 ± 0.0034 &0.6857 ± 0.0061 &0.1595 ± 0.0018 \\
    % GIN &0.7558 ± 0.0140 &0.2266 ± 0.0028 &0.6892 ± 0.0100 &0.1495 ± 0.0023 \\
    GIN+virtual node &0.7707 ± 0.0149 &0.2703 ± 0.0023 &0.7037 ± 0.0107 &0.1581 ± 0.0026 \\
    GatedGCN-LSPE &-- &0.267 ± 0.002 &-- &-- \\
    PNA &0.7905 ± 0.0132 &0.2838 ± 0.0035 &-- &0.1570 ± 0.0032 \\
    DeeperGCN &0.7858 ± 0.0117 &0.2781 ± 0.0038 &\third{0.7712 ± 0.0071} &-- \\
    % DAGNN & &-- &-- &0.1751 ± 0.0049 \\
    DGN &\third{0.7970 ± 0.0097} &0.2885 ± 0.0030 &-- &-- \\
    GSN (directional) &\second{0.8039 ± 0.0090} &-- &-- &-- \\
    GSN (GIN+VN base) &0.7799 ± 0.0100 &-- &-- &-- \\
    CIN &\first{0.8094 ± 0.0057} &-- &-- &-- \\
    GIN-AK+ &0.7961 ± 0.0119 &\second{0.2930 ± 0.0044} &-- &-- \\
    CRaWl &-- &\first{0.2986 ± 0.0025} &-- &-- \\
    ExpC~\cite{yang2022ExpC} &0.7799 ± 0.0082 &0.2342 ± 0.0029 &\second{0.7976 ± 0.0072} &-- \\\midrule
    SAN &0.7785 ± 0.2470 &0.2765 ± 0.0042 &-- &-- \\
    % Graphormer (pre-trained) &0.8051 ± 0.0053 &0.3140 ± 0.0032 &-- &-- \\
    GraphTrans (GCN-Virtual) &-- &0.2761 ± 0.0029 &-- &\third{0.1830 ± 0.0024} \\
    K-Subtree SAT &-- &-- &0.7522 ± 0.0056 &\first{0.1937 ± 0.0028} \\\midrule
    \method (ours) &0.7880 ± 0.0101 &\third{0.2907 ± 0.0028} &\first{0.8015 ± 0.0033} &\second{0.1894 ± 0.0024} \\
    \bottomrule
    \end{tabular}
\end{table}

\begin{table}
    % \vspace{-10pt}
    \caption{Evaluation on PCQM4Mv2~\cite{hu2021ogblsc} dataset.
    % Shown are results of a single run as is the standard for this dataset.
    For \method evaluation, we treated the \emph{validation} set of the dataset as a test set, since the \emph{test-dev} set labels are private.
    For more details refer to Appendix~\ref{app:exp_details}.
    }
    \label{tab:results_pcqm4m}
    \centering
    % \small
    \fontsize{8.5pt}{8.5pt}\selectfont
    \begin{tabular}{lccccc}\toprule
    \multirow{2}{*}{\textbf{Model}} &\multicolumn{3}{c}{\textbf{PCQM4Mv2}} \\\cmidrule{2-5}
    &\textbf{Test-dev MAE $\downarrow$} &\textbf{Validation MAE $\downarrow$} &\textbf{Training MAE} &\textbf{\# Param.} \\\midrule
    GCN &0.1398 &0.1379 & n/a &2.0M \\
    GCN-virtual &0.1152 &0.1153 & n/a &4.9M \\
    GIN &0.1218 &0.1195 & n/a &3.8M \\
    GIN-virtual &0.1084 &0.1083 & n/a &6.7M \\\midrule
    GRPE~\cite{park2022GRPE} &0.0898 &0.0890 & n/a &46.2M \\
    EGT~\cite{hussain2022EGT} &0.0872 &\third{0.0869} & n/a &89.3M \\
    Graphormer~\cite{shi2022benchgraphormer} &n/a &\second{0.0864} &0.0348 & 48.3M \\\midrule
    \method-small &n/a &0.0938 &0.0653 &6.2M \\
    \method-medium &n/a &\first{0.0858} &0.0726  &19.4M \\
    \bottomrule
    \end{tabular}
    \vspace{-5pt}
    % \vspace{5pt}
\end{table}

% \begin{wraptable}{r}{0.35\textwidth}
% % \begin{table}
%     \caption{MalNet-Tiny of 5k graphs of up to 5k nodes each~\cite{freitas2021malnet}.}
%     \label{tab:results_malnet}
%     \centering
%     \small
%     \begin{tabular}{lcc}\toprule
%     \multirow{2}{*}{\textbf{Model}} &\textbf{MalNet-Tiny} \\
%     &\textbf{Accuracy $\uparrow$} \\\midrule
%     GIN &90\% \\
%     GCN &81\% \\\midrule
%     % GCN-JK~\cite{lo2022malware} &89.7\% \\
%     % GraphSAGE-JK~\cite{lo2022malware} &94.4\% \\
%     % GIN-JK~\cite{lo2022malware} &90.0\% \\\midrule
%     % \method &93.2\% ± 0.283 \\  %<-- 6layer 128dim 3repeats 
%     \method (w/Perf.) &92.72\% ± 0.7pp \\  %<-- 5layer 64dim 10repeats 
%     \method (w/Transf.) &93.36\% ± 0.6pp \\  %<-- 5layer 64dim 10repeats 
%     \bottomrule
%     \end{tabular}
% % \end{table}
% \end{wraptable}

\myparagraph{Long-Range Graph Benchmark~\cite{dwivedi2022LRGB}.}
Finally, we evaluate the GPS method on a recent Long-Range Graph Benchmark (LRGB) suite of 5 datasets that are intended to test a method's ability to capture long-range dependencies in the input graphs. We abide to the \textasciitilde500k model parameter budget and closely follow the experimental setup and hyperparameter choices of the graph Transformer baselines tested in LRGB~\cite{dwivedi2022LRGB}. We keep the same node/edge encoders and model depth (number of layers), deviating only in two aspects: i) we slightly decrease the size of hidden node representations to fit within the parameter budget, ii) we employ cosine learning rate schedule as in our other experiments (Section \ref{app:hyperparams}). For each dataset we utilize LapPE positional encodings and GPS with GatedGCN~\cite{bresson2017GatedGCN} and Transformer~\cite{vaswani2017attention} components.

GPS improves over all evaluated baselines in 4 out of 5 LRGB datasets (Table~\ref{tab:results_lrgb}). Additionally, we conducted GPS ablation studies on PascalVOC-SP and Peptides-func datasets in the same fashion as for 4 previous datasets in Table~\ref{fig:ablations}, presented in Tables~\ref{tab:abl_pascal}~and~\ref{tab:abl_peptidesfunc}, respectively. For both datasets the global attention, in form of Transformer or Performer, is shown to be a critical component of the GPS in outperforming MPNNs. In the case of PascalVOC-SP, interestingly, the Laplacian PEs are not beneficial, as without them the GPS scores even higher F1-score $0.3846$, and PEG~\cite{wang2022equivstable} relative distance embeddings enable the highest score of $0.3956$.

\begin{table}[t]
\caption{Test performance on long-range graph benchmarks (LRGB)~\cite{dwivedi2022LRGB}. Shown is the mean~±~s.d.~of 4 runs. The \first{first}, \second{second}, and \third{third} best are highlighted. \\ $^*$SAN on COCO-SP exceeded 60h time limit on an NVidia A100 GPU system.}
    \label{tab:results_lrgb}
    \fontsize{8.5pt}{8.5pt}\selectfont
    \setlength\tabcolsep{4pt} % default value: 6pt
    \begin{adjustwidth}{-2.5 cm}{-2.5 cm}\centering
    \begin{tabular}{lccccc}\toprule
    \multirow{2}{*}{\textbf{Model}} &\textbf{PascalVOC-SP} &\textbf{COCO-SP} &\textbf{Peptides-func} &\textbf{Peptides-struct} &\textbf{PCQM-Contact} \\\cmidrule{2-6}
    &\textbf{F1 score $\uparrow$} &\textbf{F1 score $\uparrow$} &\textbf{AP $\uparrow$} &\textbf{MAE $\downarrow$} &\textbf{MRR $\uparrow$} \\\midrule
    GCN &0.1268 ± 0.0060 &0.0841 ± 0.0010 &0.5930 ± 0.0023 &0.3496 ± 0.0013 &0.3234 ± 0.0006 \\
    GINE &0.1265 ± 0.0076 &0.1339 ± 0.0044 &0.5498 ± 0.0079 &0.3547 ± 0.0045 &0.3180 ± 0.0027 \\
    GatedGCN &0.2873 ± 0.0219 &\second{0.2641 ± 0.0045} &0.5864 ± 0.0077 &0.3420 ± 0.0013 &0.3218 ± 0.0011 \\
    GatedGCN+RWSE &0.2860 ± 0.0085 &0.2574 ± 0.0034 &0.6069 ± 0.0035 &0.3357 ± 0.0006 &0.3242 ± 0.0008 \\ \midrule
    Transformer+LapPE &0.2694 ± 0.0098 &\third{0.2618 ± 0.0031} &0.6326 ± 0.0126 &\second{0.2529 ± 0.0016} &0.3174 ± 0.0020 \\
    SAN+LapPE &\second{0.3230 ± 0.0039} &0.2592 ± 0.0158* &\third{0.6384 ± 0.0121} &0.2683 ± 0.0043 &\first{0.3350 ± 0.0003} \\
    SAN+RWSE &\third{0.3216 ± 0.0027} &0.2434 ± 0.0156* &\second{0.6439 ± 0.0075} &\third{0.2545 ± 0.0012} &\second{0.3341 ± 0.0006} \\ \midrule
    GPS (ours) &\first{0.3748 ± 0.0109} &\first{0.3412 ± 0.0044} &\first{0.6535 ± 0.0041} &\first{0.2500 ± 0.0005} &\third{0.3337 ± 0.0006} \\
    \bottomrule
    \end{tabular}
    \end{adjustwidth}
\end{table}

%##############################################################################
\vspace{-3pt}
% \newpage
\section{Conclusion}\label{sec:conclusion}
\vspace{-3pt}
Our work is setting the foundation for a unified architecture of graph neural networks, with modular and scalable graph Transformers and a broader understanding of the role of graphs with positional and structural encodings. In our ablation studies, we demonstrated the importance of each module: the Transformer, flexible message-passing, and rich positional and structural encodings all contributed to the success of \method on a wide variety of benchmarks. Indeed, considering 5 Benchmarking-GNN tasks \cite{dwivedi2020benchmarking}, 5 OGB(-LSC) tasks \cite{hu2020ogb,hu2021ogblsc}, 5 LRGB tasks \cite{dwivedi2022LRGB} and MalNet-Tiny, we outperformed every graph Transformer on 11 out of 16 tasks while also achieving state-of-the-art on 8 of them. We further showed that the model can scale to very large graphs of several thousand nodes, far beyond any previous graph Transformer. By open-sourcing the \gtgym package, we hope to accelerate the research in efficient and expressive graph Transformers, and move the field closer to a unified hybrid Transformer architecture for graphs.

\textbf{Limitations.} We find that graph transformers are sensitive to hyperparameters and there is no \emph{one-size-fits-all} solution for all datasets. We also identify a lack of challenging graph datasets necessitating long-range dependencies where linear attention architectures could exhibit all scalability benefits. 

\textbf{Societal Impact.} As a general graph representation learning method, we do not foresee immediate negative societal outcomes. However, its particular application, e.g., in drug discovery or computational biology, will have to be thoroughly examined for trustworthiness or malicious usage.

%###########################################################
% Do NOT include Acknowledgments in the anonymized submission, only in the final paper. You can use the \texttt{ack} environment provided in the style file to automatically hide this section in the anonymized submission.
%###########################################################
\begin{ack}
This work was partially funded by IVADO (Institut de valorisation des données) grant PRF-2019-3583139727 and Canada CIFAR AI Chair~[\emph{G.W.}]. This research is supported by Nanyang Technological University, under SUG Grant (020724-00001)~[\emph{V.P.D.}] and Samsung AI graph at Mila~[\emph{M.G.}]. The content provided here is solely the responsibility of the authors and does not necessarily represent the official views of the funding agencies.
\end{ack}

%###########################################################
\bibliography{references}

\begin{thebibliography}{68}
\providecommand{\natexlab}[1]{#1}
\providecommand{\url}[1]{\texttt{#1}}
\expandafter\ifx\csname urlstyle\endcsname\relax
  \providecommand{\doi}[1]{doi: #1}\else
  \providecommand{\doi}{doi: \begingroup \urlstyle{rm}\Url}\fi

\bibitem[Alon and Yahav(2021)]{alon2021on}
Uri Alon and Eran Yahav.
\newblock On the bottleneck of graph neural networks and its practical
  implications.
\newblock In \emph{International Conference on Learning Representations}, 2021.

\bibitem[Andres et~al.(2016)Andres, Deuschel, and Slowik]{andres2016heat}
Sebastian Andres, Jean-Dominique Deuschel, and Martin Slowik.
\newblock Heat kernel estimates for random walks with degenerate weights.
\newblock \emph{Electronic Journal of Probability}, 21:\penalty0 1--21, 2016.

\bibitem[Beaini et~al.(2021)Beaini, Passaro, L{\'e}tourneau, Hamilton, Corso,
  and Li{\`o}]{beaini2021directional_dgn}
Dominique Beaini, Saro Passaro, Vincent L{\'e}tourneau, Will Hamilton, Gabriele
  Corso, and Pietro Li{\`o}.
\newblock Directional graph networks.
\newblock In \emph{International Conference on Machine Learning}, pages
  748--758. PMLR, 2021.

\bibitem[Beltagy et~al.(2020)Beltagy, Peters, and
  Cohan]{DBLP:journals/corr/abs-2004-05150}
Iz~Beltagy, Matthew~E. Peters, and Arman Cohan.
\newblock Longformer: The long-document transformer.
\newblock \emph{CoRR}, abs/2004.05150, 2020.

\bibitem[Bodnar et~al.(2021)Bodnar, Frasca, Otter, Wang, Lio, Montufar, and
  Bronstein]{bodnar2021weisfeiler_CIN}
Cristian Bodnar, Fabrizio Frasca, Nina Otter, Yuguang Wang, Pietro Lio, Guido~F
  Montufar, and Michael Bronstein.
\newblock Weisfeiler and {L}ehman go cellular: {CW} networks.
\newblock \emph{Advances in Neural Information Processing Systems},
  34:\penalty0 2625--2640, 2021.

\bibitem[Bouritsas et~al.(2022)Bouritsas, Frasca, Zafeiriou, and
  Bronstein]{bouritsas2022improving_GSN}
Giorgos Bouritsas, Fabrizio Frasca, Stefanos~P Zafeiriou, and Michael
  Bronstein.
\newblock Improving graph neural network expressivity via subgraph isomorphism
  counting.
\newblock \emph{IEEE Transactions on Pattern Analysis and Machine
  Intelligence}, 2022.

\bibitem[Bresson and Laurent(2017)]{bresson2017GatedGCN}
Xavier Bresson and Thomas Laurent.
\newblock {Residual Gated Graph ConvNets}.
\newblock \emph{arXiv:1711.07553}, 2017.

\bibitem[Chelombiev et~al.(2021)Chelombiev, Justus, Orr, Dietrich, Gressmann,
  Koliousis, and Luschi]{chelombiev2021groupbert}
Ivan Chelombiev, Daniel Justus, Douglas Orr, Anastasia Dietrich, Frithjof
  Gressmann, Alexandros Koliousis, and Carlo Luschi.
\newblock Groupbert: Enhanced transformer architecture with efficient grouped
  structures.
\newblock \emph{arXiv:2106.05822}, 2021.

\bibitem[Chen et~al.(2022)Chen, O'Bray, and Borgwardt]{chen2022SAT}
Dexiong Chen, Leslie O'Bray, and Karsten Borgwardt.
\newblock Structure-aware transformer for graph representation learning.
\newblock \emph{Proceedings of the 39th International Conference on Machine
  Learning}, 2022.

\bibitem[Chen et~al.(2019)Chen, Chen, Villar, and Bruna]{chen2019equivalence}
Zhengdao Chen, Lei Chen, Soledad Villar, and Joan Bruna.
\newblock On the equivalence between graph isomorphism testing and function
  approximation with gnns.
\newblock \emph{Advances in Neural Information Processing Systems}, 2019.

\bibitem[Choromanski et~al.(2021{\natexlab{a}})Choromanski, Lin, Chen, Zhang,
  Sehanobish, Likhosherstov, Parker-Holder, Sarlos, Weller, and
  Weingarten]{choromanski2021blocktoeplitz}
Krzysztof Choromanski, Han Lin, Haoxian Chen, Tianyi Zhang, Arijit Sehanobish,
  Valerii Likhosherstov, Jack Parker-Holder, Tamas Sarlos, Adrian Weller, and
  Thomas Weingarten.
\newblock From block-{T}oeplitz matrices to differential equations on graphs:
  towards a general theory for scalable masked transformers.
\newblock \emph{arXiv:2107.07999}, 2021{\natexlab{a}}.

\bibitem[Choromanski et~al.(2021{\natexlab{b}})Choromanski, Likhosherstov,
  Dohan, Song, Gane, Sarl{\'{o}}s, Hawkins, Davis, Mohiuddin, Kaiser, Belanger,
  Colwell, and Weller]{DBLP:conf/iclr/ChoromanskiLDSG21}
Krzysztof~Marcin Choromanski, Valerii Likhosherstov, David Dohan, Xingyou Song,
  Andreea Gane, Tam{\'{a}}s Sarl{\'{o}}s, Peter Hawkins, Jared~Quincy Davis,
  Afroz Mohiuddin, Lukasz Kaiser, David~Benjamin Belanger, Lucy~J. Colwell, and
  Adrian Weller.
\newblock Rethinking attention with performers.
\newblock In \emph{9th International Conference on Learning Representations},
  2021{\natexlab{b}}.

\bibitem[Corso et~al.(2020)Corso, Cavalleri, Beaini, Li{\`o}, and
  Veli{\v{c}}kovi{\'c}]{corso2020principal_pna}
Gabriele Corso, Luca Cavalleri, Dominique Beaini, Pietro Li{\`o}, and Petar
  Veli{\v{c}}kovi{\'c}.
\newblock Principal neighbourhood aggregation for graph nets.
\newblock \emph{Advances in Neural Information Processing Systems},
  33:\penalty0 13260--13271, 2020.

\bibitem[Dwivedi and Bresson(2020)]{dwivedi2020generalization}
Vijay~Prakash Dwivedi and Xavier Bresson.
\newblock A generalization of transformer networks to graphs.
\newblock \emph{arXiv:2012.09699}, 2020.

\bibitem[Dwivedi et~al.(2020)Dwivedi, Joshi, Laurent, Bengio, and
  Bresson]{dwivedi2020benchmarking}
Vijay~Prakash Dwivedi, Chaitanya~K Joshi, Thomas Laurent, Yoshua Bengio, and
  Xavier Bresson.
\newblock Benchmarking graph neural networks.
\newblock \emph{arXiv:2003.00982}, 2020.

\bibitem[Dwivedi et~al.(2022{\natexlab{a}})Dwivedi, Luu, Laurent, Bengio, and
  Bresson]{dwivedi2022LPE}
Vijay~Prakash Dwivedi, Anh~Tuan Luu, Thomas Laurent, Yoshua Bengio, and Xavier
  Bresson.
\newblock Graph neural networks with learnable structural and positional
  representations.
\newblock In \emph{International Conference on Learning Representations},
  2022{\natexlab{a}}.

\bibitem[Dwivedi et~al.(2022{\natexlab{b}})Dwivedi, Rampášek, Galkin, Parviz,
  Wolf, Luu, and Beaini]{dwivedi2022LRGB}
Vijay~Prakash Dwivedi, Ladislav Rampášek, Mikhail Galkin, Ali Parviz, Guy
  Wolf, Anh~Tuan Luu, and Dominique Beaini.
\newblock Long range graph benchmark.
\newblock \emph{Neural Information Processing Systems (NeurIPS 2022), Track on
  Datasets and Benchmarks}, 2022{\natexlab{b}}.

\bibitem[d’Ascoli et~al.(2021)d’Ascoli, Touvron, Leavitt, Morcos, Biroli,
  and Sagun]{dascoli2021convit}
St{\'e}phane d’Ascoli, Hugo Touvron, Matthew~L Leavitt, Ari~S Morcos, Giulio
  Biroli, and Levent Sagun.
\newblock Convit: Improving vision transformers with soft convolutional
  inductive biases.
\newblock In \emph{International Conference on Machine Learning}, pages
  2286--2296. PMLR, 2021.

\bibitem[Ertl and Schuffenhauer(2009)]{ertl2009estimation_sascore}
Peter Ertl and Ansgar Schuffenhauer.
\newblock Estimation of synthetic accessibility score of drug-like molecules
  based on molecular complexity and fragment contributions.
\newblock \emph{Journal of cheminformatics}, 1\penalty0 (1):\penalty0 1--11,
  2009.

\bibitem[Fey and Lenssen(2019)]{FeyLenssen2019PyG}
Matthias Fey and Jan~Eric Lenssen.
\newblock Fast graph representation learning with {PyTorch Geometric}.
\newblock In \emph{ICLR Workshop on Representation Learning on Graphs and
  Manifolds}, 2019.

\bibitem[Freitas et~al.(2021)Freitas, Dong, Neil, and Chau]{freitas2021malnet}
Scott Freitas, Yuxiao Dong, Joshua Neil, and Duen~Horng Chau.
\newblock A large-scale database for graph representation learning.
\newblock In \emph{35th Conference on Neural Information Processing Systems:
  Datasets and Benchmarks Track}, 2021.

\bibitem[Gilmer et~al.(2017)Gilmer, Schoenholz, Riley, Vinyals, and
  Dahl]{gilmer2017neural}
Justin Gilmer, Samuel~S Schoenholz, Patrick~F Riley, Oriol Vinyals, and
  George~E Dahl.
\newblock Neural message passing for quantum chemistry.
\newblock In \emph{International conference on machine learning}, pages
  1263--1272. PMLR, 2017.

\bibitem[Gu et~al.(2022)Gu, Goel, and Re]{gu2022efficiently}
Albert Gu, Karan Goel, and Christopher Re.
\newblock Efficiently modeling long sequences with structured state spaces.
\newblock In \emph{International Conference on Learning Representations}, 2022.

\bibitem[Guo et~al.(2021)Guo, Han, Wu, Xu, Tang, Xu, and Wang]{guo2021cmt}
Jianyuan Guo, Kai Han, Han Wu, Chang Xu, Yehui Tang, Chunjing Xu, and Yunhe
  Wang.
\newblock {CMT}: Convolutional neural networks meet vision transformers.
\newblock \emph{arXiv:2107.06263}, 2021.

\bibitem[Han et~al.(2022)Han, Wang, Chen, Chen, Guo, Liu, Tang, Xiao, Xu, Xu,
  et~al.]{han2022survey_vision_transformer}
Kai Han, Yunhe Wang, Hanting Chen, Xinghao Chen, Jianyuan Guo, Zhenhua Liu,
  Yehui Tang, An~Xiao, Chunjing Xu, Yixing Xu, et~al.
\newblock A survey on vision transformer.
\newblock \emph{IEEE Transactions on Pattern Analysis and Machine
  Intelligence}, 2022.

\bibitem[Hu et~al.(2019)Hu, Liu, Gomes, Zitnik, Liang, Pande, and
  Leskovec]{hu2019strategies_GINE}
Weihua Hu, Bowen Liu, Joseph Gomes, Marinka Zitnik, Percy Liang, Vijay Pande,
  and Jure Leskovec.
\newblock Strategies for pre-training graph neural networks.
\newblock \emph{arXiv:1905.12265}, 2019.

\bibitem[Hu et~al.(2020)Hu, Fey, Zitnik, Dong, Ren, Liu, Catasta, and
  Leskovec]{hu2020ogb}
Weihua Hu, Matthias Fey, Marinka Zitnik, Yuxiao Dong, Hongyu Ren, Bowen Liu,
  Michele Catasta, and Jure Leskovec.
\newblock Open {Graph} {Benchmark}: {Datasets} for {Machine} {Learning} on
  {Graphs}.
\newblock \emph{34th Conference on Neural Information Processing Systems},
  2020.

\bibitem[Hu et~al.(2021)Hu, Fey, Ren, Nakata, Dong, and Leskovec]{hu2021ogblsc}
Weihua Hu, Matthias Fey, Hongyu Ren, Maho Nakata, Yuxiao Dong, and Jure
  Leskovec.
\newblock {OGB}-{LSC}: A large-scale challenge for machine learning on graphs.
\newblock In \emph{35th Conference on Neural Information Processing Systems:
  Datasets and Benchmarks Track}, 2021.

\bibitem[Hussain et~al.(2022)Hussain, Zaki, and Subramanian]{hussain2022EGT}
Md~Shamim Hussain, Mohammed~J Zaki, and Dharmashankar Subramanian.
\newblock Global self-attention as a replacement for graph convolution.
\newblock In \emph{Proceedings of the 28th ACM SIGKDD Conference on Knowledge
  Discovery and Data Mining}, pages 655--665, 2022.

\bibitem[Ioffe and Szegedy(2015)]{ioffe2015batchnorm}
Sergey Ioffe and Christian Szegedy.
\newblock Batch normalization: Accelerating deep network training by reducing
  internal covariate shift.
\newblock In \emph{International conference on machine learning}, pages
  448--456. PMLR, 2015.

\bibitem[Jain et~al.(2021)Jain, Wu, Wright, Mirhoseini, Gonzalez, and
  Stoica]{jain2021graphtrans}
Paras Jain, Zhanghao Wu, Matthew Wright, Azalia Mirhoseini, Joseph~E Gonzalez,
  and Ion Stoica.
\newblock Representing long-range context for graph neural networks with global
  attention.
\newblock \emph{Advances in Neural Information Processing Systems}, 34, 2021.

\bibitem[Kalyan et~al.(2021)Kalyan, Rajasekharan, and
  Sangeetha]{kalyan2021ammus_transformer_survey}
Katikapalli~Subramanyam Kalyan, Ajit Rajasekharan, and Sivanesan Sangeetha.
\newblock Ammus: A survey of transformer-based pretrained models in natural
  language processing.
\newblock \emph{arXiv:2108.05542}, 2021.

\bibitem[Kipf and Welling(2016)]{kipf2016semi}
Thomas~N Kipf and Max Welling.
\newblock Semi-supervised classification with graph convolutional networks.
\newblock \emph{arXiv:1609.02907}, 2016.

\bibitem[Kitaev et~al.(2020)Kitaev, Kaiser, and
  Levskaya]{DBLP:conf/iclr/KitaevKL20}
Nikita Kitaev, Lukasz Kaiser, and Anselm Levskaya.
\newblock Reformer: The efficient transformer.
\newblock In \emph{8th International Conference on Learning Representations},
  2020.

\bibitem[Koutis and Le(2019)]{koutis2019spectral}
Ioannis Koutis and Huong Le.
\newblock Spectral modification of graphs for improved spectral clustering.
\newblock \emph{Advances in Neural Information Processing Systems}, 32, 2019.

\bibitem[Kreuzer et~al.(2021)Kreuzer, Beaini, Hamilton, L{\'e}tourneau, and
  Tossou]{kreuzer2021rethinking}
Devin Kreuzer, Dominique Beaini, William~L. Hamilton, Vincent L{\'e}tourneau,
  and Prudencio Tossou.
\newblock Rethinking graph transformers with spectral attention.
\newblock In \emph{Advances in Neural Information Processing Systems}, 2021.

\bibitem[Kurin et~al.(2020)Kurin, Igl, Rockt{\"a}schel, Boehmer, and
  Whiteson]{kurin2020my_body}
Vitaly Kurin, Maximilian Igl, Tim Rockt{\"a}schel, Wendelin Boehmer, and Shimon
  Whiteson.
\newblock My body is a cage: the role of morphology in graph-based incompatible
  control.
\newblock \emph{arXiv:2010.01856}, 2020.

\bibitem[Li et~al.(2020)Li, Wang, Wang, and Leskovec]{li2020distance}
Pan Li, Yanbang Wang, Hongwei Wang, and Jure Leskovec.
\newblock Distance encoding: Design provably more powerful neural networks for
  graph representation learning.
\newblock \emph{Advances in Neural Information Processing Systems},
  33:\penalty0 4465--4478, 2020.

\bibitem[Lim et~al.(2022)Lim, Robinson, Zhao, Smidt, Sra, Maron, and
  Jegelka]{lim2022sign}
Derek Lim, Joshua Robinson, Lingxiao Zhao, Tess Smidt, Suvrit Sra, Haggai
  Maron, and Stefanie Jegelka.
\newblock Sign and basis invariant networks for spectral graph representation
  learning.
\newblock \emph{arXiv:2202.13013}, 2022.

\bibitem[Liu et~al.(2022)Liu, Yu, Liao, Li, Lin, Liu, and
  Dustdar]{liu2022pyraformer}
Shizhan Liu, Hang Yu, Cong Liao, Jianguo Li, Weiyao Lin, Alex~X. Liu, and
  Schahram Dustdar.
\newblock Pyraformer: Low-complexity pyramidal attention for long-range time
  series modeling and forecasting.
\newblock In \emph{International Conference on Learning Representations}, 2022.

\bibitem[Loshchilov and Hutter(2019)]{loshchilov2018decoupled}
Ilya Loshchilov and Frank Hutter.
\newblock Decoupled weight decay regularization.
\newblock In \emph{International Conference on Learning Representations}, 2019.

\bibitem[Loukas(2020)]{loukas2020graph}
Andreas Loukas.
\newblock What graph neural networks cannot learn: depth vs width.
\newblock In \emph{International Conference on Learning Representations}, 2020.

\bibitem[Maron et~al.(2019)Maron, Ben-Hamu, Serviansky, and
  Lipman]{maron2019provably}
Haggai Maron, Heli Ben-Hamu, Hadar Serviansky, and Yaron Lipman.
\newblock Provably powerful graph networks.
\newblock \emph{arXiv:1905.11136}, 2019.

\bibitem[Mialon et~al.(2021)Mialon, Chen, Selosse, and
  Mairal]{mialon2021graphit}
Gr{\'e}goire Mialon, Dexiong Chen, Margot Selosse, and Julien Mairal.
\newblock {GraphiT}: Encoding graph structure in transformers.
\newblock \emph{arXiv:2106.05667}, 2021.

\bibitem[Morris et~al.(2019)Morris, Ritzert, Fey, Hamilton, Lenssen, Rattan,
  and Grohe]{morris2019}
Christopher Morris, Martin Ritzert, Matthias Fey, William~L. Hamilton, Jan~Eric
  Lenssen, Gaurav Rattan, and Martin Grohe.
\newblock Weisfeiler and {L}eman go neural: Higher-order graph neural networks.
\newblock In \emph{The Thirty-Third {AAAI} Conference on Artificial
  Intelligence}, pages 4602--4609. {AAAI} Press, 2019.

\bibitem[Murphy et~al.(2019)Murphy, Srinivasan, Rao, and
  Ribeiro]{murphy2019relational}
Ryan Murphy, Balasubramaniam Srinivasan, Vinayak Rao, and Bruno Ribeiro.
\newblock Relational pooling for graph representations.
\newblock In \emph{International Conference on Machine Learning}, pages
  4663--4673. PMLR, 2019.

\bibitem[Oono and Suzuki(2020)]{Oono2020Graph}
Kenta Oono and Taiji Suzuki.
\newblock Graph neural networks exponentially lose expressive power for node
  classification.
\newblock In \emph{International Conference on Learning Representations}, 2020.

\bibitem[Park et~al.(2022)Park, Chang, Lee, Kim, and won Hwang]{park2022GRPE}
Wonpyo Park, Woonggi Chang, Donggeon Lee, Juntae Kim, and Seung won Hwang.
\newblock {GRPE}: Relative positional encoding for graph transformer.
\newblock \emph{arXiv:22201.12787}, 2022.

\bibitem[Raffel et~al.(2020)Raffel, Shazeer, Roberts, Lee, Narang, Matena,
  Zhou, Li, and Liu]{raffel2020t5}
Colin Raffel, Noam Shazeer, Adam Roberts, Katherine Lee, Sharan Narang, Michael
  Matena, Yanqi Zhou, Wei Li, and Peter~J. Liu.
\newblock Exploring the limits of transfer learning with a unified text-to-text
  transformer.
\newblock \emph{Journal of Machine Learning Research}, 21\penalty0
  (140):\penalty0 1--67, 2020.

\bibitem[Sato(2020)]{sato2020survey}
Ryoma Sato.
\newblock A survey on the expressive power of graph neural networks.
\newblock \emph{arXiv:2003.04078}, 2020.

\bibitem[Shi et~al.(2022)Shi, Zheng, Ke, Shen, You, He, Luo, Liu, He, and
  Liu]{shi2022benchgraphormer}
Yu~Shi, Shuxin Zheng, Guolin Ke, Yifei Shen, Jiacheng You, Jiyan He, Shengjie
  Luo, Chang Liu, Di~He, and Tie-Yan Liu.
\newblock Benchmarking graphormer on large-scale molecular modeling datasets.
\newblock \emph{arXiv:2203.04810}, 2022.

\bibitem[Tay et~al.(2021)Tay, Dehghani, Abnar, Shen, Bahri, Pham, Rao, Yang,
  Ruder, and Metzler]{tay2021long}
Yi~Tay, Mostafa Dehghani, Samira Abnar, Yikang Shen, Dara Bahri, Philip Pham,
  Jinfeng Rao, Liu Yang, Sebastian Ruder, and Donald Metzler.
\newblock Long range arena: A benchmark for efficient transformers.
\newblock In \emph{International Conference on Learning Representations}, 2021.

\bibitem[Toenshoff et~al.(2021)Toenshoff, Ritzert, Wolf, and
  Grohe]{toenshoff2021CRaWl}
Jan Toenshoff, Martin Ritzert, Hinrikus Wolf, and Martin Grohe.
\newblock Graph learning with 1d convolutions on random walks.
\newblock \emph{arXiv:2102.08786}, 2021.

\bibitem[Topping et~al.(2021)Topping, Di~Giovanni, Chamberlain, Dong, and
  Bronstein]{topping2021understanding_ricci}
Jake Topping, Francesco Di~Giovanni, Benjamin~Paul Chamberlain, Xiaowen Dong,
  and Michael~M Bronstein.
\newblock Understanding over-squashing and bottlenecks on graphs via curvature.
\newblock \emph{arXiv:2111.14522}, 2021.

\bibitem[Vaswani et~al.(2017)Vaswani, Shazeer, Parmar, Uszkoreit, Jones, Gomez,
  Kaiser, and Polosukhin]{vaswani2017attention}
Ashish Vaswani, Noam Shazeer, Niki Parmar, Jakob Uszkoreit, Llion Jones,
  Aidan~N Gomez, {\L}ukasz Kaiser, and Illia Polosukhin.
\newblock Attention is all you need.
\newblock \emph{Advances in Neural Information Processing Systems}, 30, 2017.

\bibitem[Veličković et~al.(2018)Veličković, Cucurull, Casanova, Romero,
  Liò, and Bengio]{velickovic2018GAT}
Petar Veličković, Guillem Cucurull, Arantxa Casanova, Adriana Romero, Pietro
  Liò, and Yoshua Bengio.
\newblock Graph attention networks.
\newblock In \emph{International Conference on Learning Representations}, 2018.

\bibitem[Wang et~al.(2022)Wang, Yin, Zhang, and Li]{wang2022equivstable}
Haorui Wang, Haoteng Yin, Muhan Zhang, and Pan Li.
\newblock Equivariant and stable positional encoding for more powerful graph
  neural networks.
\newblock In \emph{International Conference on Learning Representations}, 2022.

\bibitem[Wang et~al.(2020)Wang, Li, Khabsa, Fang, and Ma]{Wang2020LinformerSW}
Sinong Wang, Belinda~Z. Li, Madian Khabsa, Han Fang, and Hao Ma.
\newblock Linformer: Self-attention with linear complexity.
\newblock \emph{arXiv:2006.04768}, 2020.

\bibitem[Weisfeiler and Leman(1968)]{weisfeiler1968reduction}
Boris Weisfeiler and Andrei Leman.
\newblock The reduction of a graph to canonical form and the algebra which
  appears therein.
\newblock \emph{NTI, Series}, 2\penalty0 (9):\penalty0 12--16, 1968.

\bibitem[Wildman and Crippen(1999)]{wildman1999prediction_logp}
Scott~A Wildman and Gordon~M Crippen.
\newblock Prediction of physicochemical parameters by atomic contributions.
\newblock \emph{Journal of chemical information and computer sciences},
  39\penalty0 (5):\penalty0 868--873, 1999.

\bibitem[Xu et~al.(2019)Xu, Hu, Leskovec, and Jegelka]{xu2018how}
Keyulu Xu, Weihua Hu, Jure Leskovec, and Stefanie Jegelka.
\newblock How powerful are graph neural networks?
\newblock In \emph{International Conference on Learning Representations}, 2019.

\bibitem[Yang et~al.(2022)Yang, Wang, Shen, Qi, and Yin]{yang2022ExpC}
Mingqi Yang, Renjian Wang, Yanming Shen, Heng Qi, and Baocai Yin.
\newblock Breaking the expression bottleneck of graph neural networks.
\newblock \emph{IEEE Transactions on Knowledge and Data Engineering}, 2022.

\bibitem[Ying et~al.(2021{\natexlab{a}})Ying, Cai, Luo, Zheng, Ke, He, Shen,
  and Liu]{ying2021graphormer}
Chengxuan Ying, Tianle Cai, Shengjie Luo, Shuxin Zheng, Guolin Ke, Di~He,
  Yanming Shen, and Tie-Yan Liu.
\newblock Do transformers really perform badly for graph representation?
\newblock In \emph{Advances in Neural Information Processing Systems},
  2021{\natexlab{a}}.

\bibitem[Ying et~al.(2021{\natexlab{b}})Ying, Yang, Zheng, Ke, Luo, Cai, Wu,
  Wang, Shen, and He]{ying2021first}
Chengxuan Ying, Mingqi Yang, Shuxin Zheng, Guolin Ke, Shengjie Luo, Tianle Cai,
  Chenglin Wu, Yuxin Wang, Yanming Shen, and Di~He.
\newblock First place solution of {KDD Cup} 2021 \& {OGB} large-scale challenge
  graph prediction track.
\newblock \emph{arXiv:2106.08279}, 2021{\natexlab{b}}.

\bibitem[You et~al.(2020)You, Ying, and Leskovec]{you2020design}
Jiaxuan You, Rex Ying, and Jure Leskovec.
\newblock Design space for graph neural networks.
\newblock In \emph{Advances in Neural Information Processing Systems}, 2020.

\bibitem[Zaheer et~al.(2020)Zaheer, Guruganesh, Dubey, Ainslie, Alberti,
  Onta{\~{n}}{\'{o}}n, Pham, Ravula, Wang, Yang, and
  Ahmed]{DBLP:conf/nips/ZaheerGDAAOPRWY20}
Manzil Zaheer, Guru Guruganesh, Kumar~Avinava Dubey, Joshua Ainslie, Chris
  Alberti, Santiago Onta{\~{n}}{\'{o}}n, Philip Pham, Anirudh Ravula, Qifan
  Wang, Li~Yang, and Amr Ahmed.
\newblock {Big Bird}: Transformers for longer sequences.
\newblock In \emph{Advances in Neural Information Processing Systems}, 2020.

\bibitem[Zhang et~al.(2021)Zhang, Li, Xia, Wang, and Jin]{zhang2021labeling}
Muhan Zhang, Pan Li, Yinglong Xia, Kai Wang, and Long Jin.
\newblock Labeling trick: A theory of using graph neural networks for
  multi-node representation learning.
\newblock \emph{Advances in Neural Information Processing Systems}, 34, 2021.

\bibitem[Zhao et~al.(2022)Zhao, Jin, Akoglu, and Shah]{zhao2021stars}
Lingxiao Zhao, Wei Jin, Leman Akoglu, and Neil Shah.
\newblock From stars to subgraphs: Uplifting any {GNN} with local structure
  awareness.
\newblock In \emph{International Conference on Learning Representations}, 2022.

\end{thebibliography}
\bibliographystyle{plainnat}

%%%%%%%%%%%%%%%%%%%%%%%%%%%%%%%%%%%%%%%%%%%%%%%%%%%%%%%%%%%%%%%%%%%%%%%%%%%%%%%
\newpage
\section*{Checklist}
%%%%%%%%%%%%%%%%%%%%%%%%%%%%%%%%%%%%%%%%%%%%%%%%%%%%%%%%%%%%%%%%%%%%%%%%%%%%%%%

% %%% BEGIN INSTRUCTIONS %%%
% The checklist follows the references.  Please
% read the checklist guidelines carefully for information on how to answer these
% questions.  For each question, change the default \answerTODO{} to \answerYes{},
% \answerNo{}, or \answerNA{}.  You are strongly encouraged to include a {\bf
% justification to your answer}, either by referencing the appropriate section of
% your paper or providing a brief inline description.  For example:
% \begin{itemize}
%   \item Did you include the license to the code and datasets? \answerYes{See Section~X.} %\answerYes{See Section~%\ref{gen_inst}.}
%   \item Did you include the license to the code and datasets? \answerNo{The code and the data are proprietary.}
%   \item Did you include the license to the code and datasets? \answerNA{}
% \end{itemize}
% Please do not modify the questions and only use the provided macros for your
% answers.  Note that the Checklist section does not count towards the page
% limit.  In your paper, please delete this instructions block and only keep the
% Checklist section heading above along with the questions/answers below.
% %%% END INSTRUCTIONS %%%

\begin{enumerate}

\item For all authors...
\begin{enumerate}
  \item Do the main claims made in the abstract and introduction accurately reflect the paper's contributions and scope?
    \answerYes{}
  \item Did you describe the limitations of your work?
    \answerYes{We discuss limitations in Section~\ref{sec:conclusion}}
  \item Did you discuss any potential negative societal impacts of your work?
    \answerYes{See Section~\ref{sec:conclusion}}
  \item Have you read the ethics review guidelines and ensured that your paper conforms to them?
    \answerYes{}
\end{enumerate}

\item If you are including theoretical results...
\begin{enumerate}
  \item Did you state the full set of assumptions of all theoretical results?
    \answerYes{}
        \item Did you include complete proofs of all theoretical results?
    \answerYes{We provide a sketch in Section~\ref{sec:theory} and more details in Appendix~\ref{app:theory}.}
\end{enumerate}

\item If you ran experiments...
\begin{enumerate}
  \item Did you include the code, data, and instructions needed to reproduce the main experimental results (either in the supplemental material or as a URL)?
    \answerYes{Code, data, and instructions are available in the supplementary material. We also include the performance traces/logs from our benchmarking experiments, supporting the results reported herein.}
  \item Did you specify all the training details (e.g., data splits, hyperparameters, how they were chosen)?
    \answerYes{We describe the datasets in Appendix~\ref{app:datasets}, splits in Appendix~\ref{app:splits}, hyperparameters in Appendix~\ref{app:hyperparams}. Full configuration files are provided in the supplementary material.}
        \item Did you report error bars (e.g., with respect to the random seed after running experiments multiple times)?
    \answerYes{We include standard deviations over several random seeds depending on the dataset evaluation protocol, more details are in Appendix~\ref{app:splits}.}
        \item Did you include the total amount of compute and the type of resources used (e.g., type of GPUs, internal cluster, or cloud provider)?
    \answerYes{We elaborate on the compute and used resources in Appendix~\ref{app:cluster}. }
\end{enumerate}

\item If you are using existing assets (e.g., code, data, models) or curating/releasing new assets...
\begin{enumerate}
  \item If your work uses existing assets, did you cite the creators?
    \answerYes{}
  \item Did you mention the license of the assets?
    \answerYes{For datasets see Appendix~\ref{app:datasets}, for software see Appendix~\ref{app:cluster}.}
  \item Did you include any new assets either in the supplemental material or as a URL?
    \answerYes{The source code of \gtgym is available in the supplementary material.}
  \item Did you discuss whether and how consent was obtained from people whose data you're using/curating?
    \answerNA{}
  \item Did you discuss whether the data you are using/curating contains personally identifiable information or offensive content?
    \answerNA{The used benchmarking datasets come from: the molecular domain with no personal information, anonymized source code (ogbg-code2 and MalNet-Tiny), anonymized images (MNIST, CIFAR10), or are statistically generated.}
\end{enumerate}

\item If you used crowdsourcing or conducted research with human subjects...
\begin{enumerate}
  \item Did you include the full text of instructions given to participants and screenshots, if applicable?
    \answerNA{}
  \item Did you describe any potential participant risks, with links to Institutional Review Board (IRB) approvals, if applicable?
    \answerNA{}
  \item Did you include the estimated hourly wage paid to participants and the total amount spent on participant compensation?
    \answerNA{}
\end{enumerate}

\end{enumerate}

%%%%%%%%%%%%%%%%%%%%%%%%%%%%%%%%%%%%%%%%%%%%%%%%%%%%%%%%%%%%%%%%%%%%%%%%%%%%%%%
% APPENDIX
%%%%%%%%%%%%%%%%%%%%%%%%%%%%%%%%%%%%%%%%%%%%%%%%%%%%%%%%%%%%%%%%%%%%%%%%%%%%%%%
\clearpage
\appendix
\renewcommand\thefigure{\thesection.\arabic{figure}}
\renewcommand\thetable{\thesection.\arabic{table}}

\section{Experimental Details}\label{app:exp_details}
\setcounter{figure}{0}
\setcounter{table}{0}

\subsection{Datasets description}
\label{app:datasets}
% Datasets listed here are used under the MIT License provided by \citet{dwivedi2020benchmarking,hu2020ogb,hu2021ogblsc} and the CC-BY License by \citet{freitas2021malnet}.

\begin{table}[h]
    \caption{Overview of the graph learning dataset~\cite{dwivedi2020benchmarking,hu2020ogb,hu2021ogblsc,freitas2021malnet,dwivedi2022LRGB} used in this study.}
    \label{tab:ds_summary}
    \begin{adjustwidth}{-2.5 cm}{-2.5 cm}
    \setlength\tabcolsep{4pt} % default value: 6pt
    \centering
    \footnotesize
    \begin{tabular}{lrrrccccc}\toprule
    % \textbf{Dataset} &\textbf{\# Graphs} &\textbf{Avg \# nodes} &\textbf{Avg \# edges} &\textbf{Directed} &\textbf{Prediction level} &\textbf{Prediction task} &\textbf{Metric} \\\midrule
    \multirow{2}{*}{\textbf{Dataset}} &\multirow{2}{*}{\textbf{\# Graphs}} &\textbf{Avg. \#} &\textbf{Avg. \#} &\multirow{2}{*}{\textbf{Directed}} &\textbf{Prediction} &\textbf{Prediction} &\multirow{2}{*}{\textbf{Metric}} \\
& &\textbf{nodes} &\textbf{edges} & &\textbf{level} &\textbf{task} & \\\midrule
    ZINC &12,000 &23.2 &24.9 &No &graph &regression &Mean Abs. Error \\
    MNIST &70,000 &70.6 &564.5 &Yes &graph &10-class classif. &Accuracy \\
    CIFAR10 &60,000 &117.6 &941.1 &Yes &graph &10-class classif. &Accuracy \\
    PATTERN &14,000 &118.9 &3,039.3 &No &inductive node &binary classif. &Accuracy \\
    CLUSTER &12,000 &117.2 &2,150.9 &No &inductive node &6-class classif. &Accuracy \\\midrule
    ogbg-molhiv &41,127 &25.5 &27.5 &No &graph &binary classif. &AUROC \\
    ogbg-molpcba &437,929 &26.0 &28.1 &No &graph &128-task classif. &Avg. Precision \\
    ogbg-ppa &158,100 &243.4 &2,266.1 &No &graph &37-task classif. &Accuracy \\
    ogbg-code2 &452,741 &125.2 &124.2 &Yes &graph &5 token sequence &F1 score \\
    PCQM4Mv2 &3,746,620 &14.1 &14.6 &No &graph &regression &Mean Abs. Error \\\midrule
    MalNet-Tiny &5,000 &1,410.3 &2,859.9 &Yes &graph &5-class classif. &Accuracy \\\midrule
    PascalVOC-SP & 11,355 & 479.4 & 2,710.5 & No &inductive node &21-class classif. &F1 score \\
    COCO-SP & 123,286 & 476.9 & 2,693.7 & No &inductive node &81-class classif. &F1 score \\
    PCQM-Contact & 529,434 & 30.1 & 61.0 & No &inductive link &link ranking &MRR \\
    Peptides-func & 15,535 & 150.9 & 307.3 & No &graph &10-task classif. &Avg. Precision \\
    Peptides-struct & 15,535 & 150.9 & 307.3 & No &graph &11-task regression &Mean Abs. Error \\
    \bottomrule
    \end{tabular}
\end{adjustwidth}\end{table}

\myparagraph{ZINC}~\cite{dwivedi2020benchmarking} (MIT License) consists of 12K molecular graphs from the ZINC database of commercially available chemical compounds. These molecular graphs are between 9 and 37 nodes large. Each node represents a heavy atom (28 possible atom types) and each edge represents a bond (3 possible types). The task is to regress constrained solubility (logP) of the molecule. The dataset comes with a predefined 10K/1K/1K train/validation/test split.

\myparagraph{MNIST and CIFAR10}~\cite{dwivedi2020benchmarking} (CC BY-SA 3.0 and MIT License) are derived from like-named image classification datasets by constructing an 8 nearest-neighbor graph of SLIC superpixels for each image. The 10-class classification tasks and standard dataset splits follow the original image classification datasets, i.e., for MNIST 55K/5K/10K and for CIFAR10 45K/5K/10K train/validation/test graphs.

\myparagraph{PATTERN and CLUSTER}~\cite{dwivedi2020benchmarking} (MIT License) are synthetic datasets sampled from Stochastic Block Model. Unlike other datasets, the prediction task here is an inductive node-level classification. In PATTERN the task is to recognize which nodes in a graph belong to one of 100 possible sub-graph patterns that were randomly generated with different SBM parameters than the rest of the graph. In CLUSTER, every graph is composed of 6 SBM-generated clusters, each drawn from the same distribution, with only a single node per cluster containing a unique cluster ID. The task is to infer which cluster ID each node belongs to.

\myparagraph{ogbg-molhiv and ogbg-molpcba}~\cite{hu2020ogb} (MIT License) are molecular property prediction datasets adopted by OGB from MoleculeNet. These datasets use a common node (atom) and edge (bond) featurization that represent chemophysical properties. The prediction task of ogbg-molhiv is binary classification of molecule's fitness to inhibit HIV replication. The ogbg-molpcba, derived from PubChem BioAssay, targets to predict results of 128 bioassays in multi-task binary classification setting.

\myparagraph{ogbg-ppa}~\cite{hu2020ogb} (CC-0 license) consists of protein-protein association (PPA) networks derived from 1581 species categorized to 37 taxonomic groups. Nodes represent proteins and edges encode the normalized level of 7 different associations between two proteins. The task is to classify which of the 37 groups does a PPA network originate from.

\myparagraph{ogbg-code2}~\cite{hu2020ogb} (MIT License) is comprised of abstract syntax trees (ASTs) derived from source code of functions written in Python. The task is to predict the first 5 subtokens of the original function's name.

A small number of these ASTs are much larger than the average size in the dataset. Therefore we truncated ASTs with over 1000 nodes and kept the first 1000 nodes according to their depth in the AST. This impacted 2521 (0.5\%) graphs in the dataset.

\myparagraph{OGB-LSC PCQM4Mv2}~\cite{hu2021ogblsc} (CC BY 4.0 license) is a large-scale molecular dataset that shares the same featurization as ogbg-mol* datasets. The task is to regress the HOMO-LUMO gap, a quantum physical property originally calculated using Density Functional Theory. True labels for original ``test-dev'' and ``test-challange'' dataset splits are kept private by the OGB-LSC challenge organizers. Therefore for the purpose of this paper we used the original \emph{validation} set as the test set, while we left out random 150K molecules for our validation set.

\myparagraph{PCQM4Mv2-Subset} (under the original PCQM4Mv2 CC BY 4.0 license) is a subset of PCQM4Mv2~\cite{hu2021ogblsc} that we created for the purpose of our ablation study. We sub-sampled the above-mentioned version of PCQM4Mv2 as follows; training set: 10\%; validation set: 33\%; test set: unchanged. This resulted in retaining 446,405 molecular graphs in total.

\myparagraph{MalNet-Tiny}~\cite{freitas2021malnet} (CC-BY license) is a subset of MalNet that is comprised of function call graphs (FCGs) derived from Android APKs. This subset contains 5,000 graphs of up to 5,000 nodes, each coming from benign software or 4 types of malware. The FCGs are stripped of any original node or edge features, the task is to predict the type of the software based on the structure alone. The benchmarking version of this dataset typically uses Local Degree Profile as the set of node features.

\myparagraph{PascalVOC-SP and COCO-SP}~\cite{dwivedi2022LRGB} (Custom license for Pascal VOC 2011 respecting Flickr terms of use, and CC BY 4.0 license)
are derived by SLIC superpixelization of Pascal VOC and MS COCO image datasets. Both are node classification datasets, where each superpixel node belongs to a particular object class.

\myparagraph{PCQM-Contact}~\cite{dwivedi2022LRGB} (CC BY 4.0) is derived from PCQM4Mv2 and respective 3D molecular structures. The task is a binary link prediction, identifying pairs of nodes that are considered to be in 3D contact (<3.5{\AA}) yet distant in the 2D graph (>5 hops). The default evaluation ranking metric used is the Mean Reciprocal Rank (MRR).

\myparagraph{Peptides-func and Peptides-struct}~\cite{dwivedi2022LRGB} (CC BY-NC 4.0)
are both composed of atomic graphs of peptides retrieved from SATPdb. In Peptides-func the prediction is multi-label graph classification into 10 nonexclusive peptide functional classes. While for Peptides-struct the task is graph regression of 11 3D structural properties of the peptides.

\subsection{Dataset splits and random seeds}\label{app:splits}
All evaluated benchmarks define a standard train/validation/test dataset split. We follow these and report mean performance and standard deviation from multiple execution runs with different random seeds.

All main benchmarking results are based on 10 executed runs, except PCQM4Mv2 (for which we show the result of a single random seed run) and LRGB (for which we use 4 seed).
% Results shown in Tables~\ref{tab:results_benchgnns}~and~\ref{tab:results_ogb} are based on 10 executed runs, while for MalNet-Tiny we used 3 runs and for PCQM4Mv2 we used the result of a single random seed, respectively.
The OGB-LSC~\cite{hu2021ogblsc} leaderboard for PCQM4Mv2 does not keep track of variance w.r.t. random seeds. This is likely due to the size of the dataset, in our evaluation we had run 3 random seeds and the standard deviation for \method-small was 0.00034 which is below the presentation precision.

For ablation studies we used a reduce number of 4 random seeds due to computational constraints, while for PCQM4Mv2-Subset and MalNet-Tiny we used 3 random seeds. All experiments in the ablation studies were run from scratch, results from the main text (with 10 repeats) were not reused.

\subsection{Hyperparameters}\label{app:hyperparams}
In our hyperparameter search, we experimented with a variety of positional and structural encodings, MPNN types, global attention mechanisms and their hyperparameters. Considering the large number of hyperparameters and datasets, we did not perform an exhaustive search or a grid search beyond the ablation studies presented in the main text, Section~\ref{sec:ablations}. We have extrapolated from those results and established the PE/SE type and layer types for the remaining datasets. For each dataset we then adjusted the number of layers, dimensionality $d^\ell$, and other remaining hyperparameters based on hyperparameters reported in the related literature, or eventually based on validation performance using ``line search'' along one of the hyperparameters at a time. Namely, we followed several hyperparameter choices of SAN~\cite{kreuzer2021rethinking}, SAT~\cite{chen2022SAT}, Graphormer~\cite{ying2021graphormer}, and \citet{freitas2021malnet}.

For benchmarking datasets from \citet{dwivedi2020benchmarking} we followed the most commonly used parameter budgets: up to 500k parameters for ZINC, PATTERN, and CLUSTER; and \textasciitilde100k parameters for MNIST and CIFAR10.

The final hyperparameters are presented in Tables~\ref{tab:hparams_benchgnns}, \ref{tab:hparams_ogbg}, \ref{tab:hparams_extra}, \ref{tab:hparams_lrgb}, together with the number of parameters and median wall-clock run-time for node encoding precomputation, one full epoch (including validation and test split evaluation), and the total time spent in the main loop. See Section~\ref{app:cluster} for more details on the run-time measurements.

In all our experiments we used AdamW~\cite{loshchilov2018decoupled} optimizer, with the default settings of $\beta_1=0.9$, $\beta_2=0.999$, and $\epsilon=10^{-8}$, together with linear ``warm-up'' increase of the learning rate at the beginning of the training followed by its cosine decay. The length of the warm-up period, base learning rate, and the total number of epoch were adjusted per dataset and are listed together with other hyperparameters (Tables~\ref{tab:hparams_benchgnns}, \ref{tab:hparams_ogbg}, \ref{tab:hparams_extra}, \ref{tab:hparams_lrgb}).

\begin{table}[ht]
    \caption{\method hyperparameters for five datasets from \citet{dwivedi2020benchmarking}.}
    \label{tab:hparams_benchgnns}
    \centering
    % \small
    \fontsize{8.5pt}{8.5pt}\selectfont
    \begin{tabular}{lcccccc}\toprule
    Hyperparameter &\textbf{ZINC} &\textbf{MNIST} &\textbf{CIFAR10} &\textbf{PATTERN} &\textbf{CLUSTER} \\\midrule
    \# \method Layers &10 &3 &3 &6 &16 \\
    Hidden dim &64 &52 &52 &64 &48 \\
    \method-MPNN &GINE &GatedGCN &GatedGCN &GatedGCN &GatedGCN \\
    \method-GlobAttn &Transformer &Transformer &Transformer &Transformer &Transformer \\
    \# Heads &4 &4 &4 &4 &8 \\
    Dropout &0 &0 &0 &0 &0.1 \\
    Attention dropout &0.5 &0.5 &0.5 &0.5 &0.5 \\
    Graph pooling &sum &mean &mean &-- &-- \\\midrule
    Positional Encoding &RWSE-20 &LapPE-8 &LapPE-8 &LapPE-16 &LapPE-10 \\
    PE dim &28 &8 &8 &16 &16 \\
    PE encoder &linear &DeepSet &DeepSet &DeepSet &DeepSet \\\midrule
    Batch size &32 &16 &16 &32 &16 \\
    Learning Rate &0.001 &0.001 &0.001 &0.0005 &0.0005 \\
    \# Epochs &2000 &100 &100 &100 &100 \\
    \# Warmup epochs &50 &5 &5 &5 &5 \\
    Weight decay &1e-5 &1e-5 &1e-5 &1e-5 &1e-5 \\\midrule
    \# Parameters &423,717 &115,394 &112,726 &337,201 &502,054 \\
    PE precompute &23s &96s &2.55min &28s &67s \\
    Time (epoch/total) &21s / 11.67h &76s / 2.13h &64s / 1.78h &32s / 0.89h &86s / 2.40h \\
    \bottomrule
    \end{tabular}
\end{table}

\begin{table}[ht]
    \caption{\method hyperparameters for graph-level prediction datasets from OGB~\cite{hu2020ogb}.}
    \label{tab:hparams_ogbg}
    \centering
    % \small
    \fontsize{8.5pt}{8.5pt}\selectfont
    \begin{tabular}{lccccc}\toprule
    Hyperparameter &\textbf{ogbg-molhiv} &\textbf{ogbg-molpcba} &\textbf{ogbg-ppa} &\textbf{ogbg-code2} \\\midrule
    \# \method Layers &10 &5 &3 &4 \\
    Hidden dim &64 &384 &256 &256 \\
    \method-MPNN &GatedGCN &GatedGCN &GatedGCN &GatedGCN \\
    \method-GlobAttn &Transformer &Transformer &Performer &Performer \\
    \# Heads &4 &4 &8 &4 \\
    Dropout &0.05 &0.2 &0.1 &0.2 \\
    Attention dropout &0.5 &0.5 &0.5 &0.5 \\
    Graph pooling &mean &mean &mean &mean \\\midrule
    Positional Encoding &RWSE-16 &RWSE-16 &None &None \\
    PE dim &16 &20 &-- &-- \\
    PE encoder &linear &linear &-- &-- \\\midrule
    Batch size &32 &512 &32 &32 \\
    Learning Rate &0.0001 &0.0005 &0.0003 &0.0001 \\
    \# Epochs &100 &100 &200 &30 \\
    \# Warmup epochs &5 &5 &10 &2 \\
    Weight decay &1e-5 &1e-5 &1e-5 &1e-5 \\\midrule
    \# Parameters &558,625 &9,744,496 &3,434,533 &12,454,066 \\
    PE precompute &58s &8.33min &-- &-- \\
    Time (epoch/total) &96s / 2.64h &196s / 5.44h &276s / 15.33h &1919s / 16h \\
    \bottomrule
    \end{tabular}
\end{table}

\begin{table}[ht]
    \caption{\method hyperparameters for large-scale graph-level prediction dataset OGB-LSC PCQM4Mv2~\cite{hu2021ogblsc} and MalNet-Tiny~\cite{freitas2021malnet}. \method-medium architecture follows several hyperparameter choices of Graphormer~\cite{ying2021graphormer}. Listed run-times were measured on a single NVidia A100 GPU system.}
    \label{tab:hparams_extra}
    \centering
    % \small
    \fontsize{8.5pt}{8.5pt}\selectfont
    \begin{tabular}{lcccc}\toprule
    \multirow{2}{*}{Hyperparameter} &\textbf{PCQM4Mv2} &\textbf{PCQM4Mv2} &\multirow{2}{*}{\textbf{MalNet-Tiny}} \\
    &\textbf{(\method-small)} &\textbf{(\method-medium)} & \\\midrule
    \# \method Layers &5 &10 &5 \\
    Hidden dim &304 &384 &64 \\
    \method-MPNN &GatedGCN &GatedGCN &GatedGCN \\
    \method-SelfAttn &Transformer &Transformer &Performer \\
    \# Heads &4 &16 &4 \\
    Dropout &0 &0.1 &0 \\
    Attention dropout &0.5 &0.1 &0.5 \\
    Graph pooling &mean &mean &max \\\midrule
    Positional Encoding &RWSE-16 &RWSE-16 &None \\
    PE dim &20 &20 &-- \\
    PE encoder &linear &linear &-- \\\midrule
    Batch size &256 &256 &16 \\
    Learning Rate &0.0005 &0.0002 &0.0005 \\
    \# Epochs &100 &150 &150 \\
    \# Warmup epochs &5 &10 &10 \\
    Weight decay &0 &0 &1.00e-5 \\\midrule
    \# Parameters &6,152,001 &19,414,641 &527,237 \\
    PE precompute &47min &51min &-- \\
    Time (epoch/total) &619s / 17.18h &1124s / 46.82h &46s / 1.92h \\
    \bottomrule
    \end{tabular}
\end{table}

\begin{table}[ht]
    \caption{\method hyperparameters for 5 datasets from Long Range Graph Benchmark (LRGB)~\cite{dwivedi2022LRGB}.}
    \label{tab:hparams_lrgb}
    \centering
    % \small
    \fontsize{8.5pt}{8.5pt}\selectfont
    \begin{tabular}{lcccccc}\toprule
    Hyperparameter &\textbf{PascalVOC-SP} &\textbf{COCO-SP} &\textbf{PCQM-Contact} &\textbf{Peptides-func} &\textbf{Peptides-struct} \\\midrule
    \# GPS Layers &4 &4 &4 &4 &4 \\
    Hidden dim &96 &96 &96 &96 &96 \\
    GPS-MPNN &GatedGCN &GatedGCN &GatedGCN &GatedGCN &GatedGCN \\
    GPS-SelfAttn &Transformer &Transformer &Transformer &Transformer &Transformer \\
    \# Heads &8 &8 &4 &4 &4 \\
    Dropout &0 &0 &0 &0 &0 \\
    Attention dropout &0.5 &0.5 &0.5 &0.5 &0.5 \\
    Graph pooling &-- &-- &-- &mean &mean \\\midrule
    Positional Encoding &LapPE-10 &LapPE-10 &LapPE-10 &LapPE-10 &LapPE-10 \\
    PE dim &16 &16 &16 &16 &16 \\
    PE encoder &DeepSet &DeepSet &DeepSet &DeepSet &DeepSet \\\midrule
    Batch size &32 &32 &256 &128 &128 \\
    Learning Rate &0.0005 &0.0005 &0.0003 &0.0003 &0.0003 \\
    \# Epochs &300 &300 &200 &200 &200 \\
    \# Warmup epochs &10 &10 &10 &5 &5 \\
    Weight decay &0 &0 &0 &0 &0 \\\midrule
    \# Parameters &510,453 &516,273 &512,704 &504,362 &504,459 \\
    PE precompute &8.7min &1h 34min &5.23min &73s &73s \\
    Time (epoch/total) &17.5s / 1.46h &213s / 17.8h &154s / 8.54h &6.36s / 0.35h &6.15s / 0.34h \\
    \bottomrule
    \end{tabular}
\end{table}

\subsection{Computing environment and used resources}\label{app:cluster}
Our implementation is based on PyG and its GraphGym module \cite{FeyLenssen2019PyG,you2020design} that are provided under MIT License. All experiments were run in a shared computing cluster environment with varying CPU and GPU architectures. These involved a mix of NVidia V100 (32GB), RTX8000 (48GB), and A100 (40GB) GPUs. The resource budget for each experiment was 1 GPU, between 4 and 6 CPUs, and up to 32GB system RAM. The only exception are ogbg-ppa and PCQM4Mv2 that due to their size required up to 48GB system RAM.

To measure the run-time we used Python \verb|time.perf_counter()| function. Due to the variation in computing infrastructure and load on shared resources the execution time occasionally notably varied. Therefore for our ablation studies we used only compute nodes with NVidia A100 GPUs, which considerably improved the run-time consistency. We list the wall-clock run-time that is approximately a median of the observed durations.

%%%%%%%%%%%%%%%%%%%%%%%%%%%%%%%%%%%%%%%%%%%%%%%%%%%%%%%%%%%%%%%%%%%%%%%%%%%%%%%
% \clearpage
\newpage
\section{Detailed ablation studies}\label{app:exp_ablations}
\setcounter{figure}{0}
\setcounter{table}{0}
%%%%%%%%%%%%%%%%%%%%%%%%%%%%%%%%%%%%%%%%%%%%%%%%%%%%%%%%%%%%%%%%%%%%%%%%%%%%%%%
Here we present the detailed ablation studies on impact of various MPNN, self attention, and positional / structural encoding types on \method performance and run-time. In each case, we varied a single part of the model at a time, keeping the rest of the \method hyperparameters unchanged from the best selected architecture for a given dataset. Results on ZINC are shown in Table~\ref{tab:abl_zinc}, on PCQM4Mv2-Subset in Table~\ref{tab:abl_pcqm}, on MalNet-Tiny in Table~\ref{tab:abl_malnet}, on CIFAR10 in Table~\ref{tab:abl_cifar10}, on PascalVOC-SP in Table~\ref{tab:abl_pascal}, and on Peptides-func in Table~\ref{tab:abl_peptidesfunc}. The first data row of each table reproduces results of the best selected architecture with hyperparameters detailed in Appendix~\ref{app:exp_details}; any deviations compared to the main benchmarking results of Section~\ref{sec:benchmarking} are well within the reported standard deviation. While for benchmarking results we used 10 different random seeds, here we reduced the count due to computational cost to 4 for ZINC and CIFAR10, and 3 for PCQM4Mv2-Subset and MalNet-Tiny. All time measurements reported in this section are obtained on a system with identical hardware configuration: 1x NVidia A100 (40GB) GPU and allocation of 4 AMD Milan 7413 (2.65GHz) CPU cores.

\begin{table}[ht]
    \caption{\method ablation study on \textbf{ZINC} dataset.}
    \label{tab:abl_zinc}
    \centering
    % \footnotesize
    \fontsize{8.5pt}{8.5pt}\selectfont
    \begin{tabular}{ccccccc}\toprule
    \textbf{\method-MPNN} &\textbf{\method-GlobAttn} &\textbf{PE / SE type} &\textbf{Test MAE $\downarrow$} &\textbf{\# Param.} &\textbf{Epoch / Total} \\\midrule
    GINE &Transformer &RWSE-20 &0.070 ± 0.002 &423,717 &14s / 7.56h \\\midrule
    GINE &-- &RWSE-20 &0.070 ± 0.004 &257,317 &7s / 3.90h \\
    GINE &Performer &RWSE-20 &0.071 ± 0.002 &913,317 &18s / 9.85h \\
    GINE &BigBird &RWSE-20 &0.071 ± 0.002 &507,557 &38s / 21.20h \\\midrule
    -- &Transformer &RWSE-20 &0.217 ± 0.008 &340,517 &10s / 5.74h \\
    GatedGCN &Transformer &RWSE-20 &0.086 ± 0.002 &551,077 &18s / 9.86h \\
    PNA &Transformer &RWSE-20 &0.070 ± 0.003 &680,805 &17s / 9.46h \\\midrule
    GINE &Transformer &-- &0.113 ± 0.007 &423,873 &15s / 8.38h \\
    GINE &Transformer &LapPE-8 &0.116 ± 0.009 &423,833 &13s / 7.40h \\
    GINE &Transformer &SignNet\textsuperscript{MLP}-8 &0.090 ± 0.007 &486,957 &21s / 11.61h \\
    GINE &Transformer &SignNet\textsuperscript{DeepSets}-37 &0.079 ± 0.006 &497,933 &21s / 11.49h \\
    GINE &Transformer &PEG\textsuperscript{LapEig}-8 &0.936 ± 0.143 &426,379 &16s / 8.83h \\
    GatedGCN &Transformer &PEG\textsuperscript{LapEig}-8 &0.161 ± 0.006 &553,739 &20s / 11.07h \\
    \bottomrule
    \end{tabular}
\end{table}

\begin{table}[ht]
    \caption{Ablation study on \textbf{10\% subset of PCQM4Mv2} with \method-small (Appendix~\ref{app:exp_details}).}
    \label{tab:abl_pcqm}
    \centering
    % \footnotesize
    \fontsize{8.5pt}{8.5pt}\selectfont
    \begin{tabular}{ccccccc}\toprule
    \textbf{\method-MPNN} &\textbf{\method-GlobAttn} &\textbf{PE / SE type} &\textbf{Test MAE $\downarrow$} &\textbf{\# Param.} &\textbf{Epoch / Total} \\\midrule
    GatedGCN &Transformer &RWSE-16 &0.1159 ± 0.0004 &6,152,001 &61s / 1.70h \\\midrule
    GatedGCN &-- &RWSE-16 &0.1213 ± 0.0002 &4,297,601 &45s / 1.26h \\
    GatedGCN &Performer &RWSE-16 &0.1142 ± 0.0005 &5,855,601 &83s / 2.30h \\
    GatedGCN &BigBird &RWSE-16 &0.1237 ± 0.0022 &7,080,721 &137s / 3.81h \\\midrule
    -- &Transformer &RWSE-16 &0.3294 ± 0.0137 &3,827,921 &42s / 1.16h \\
    GINE &Transformer &RWSE-16 &0.1284 ± 0.0037 &4,755,121 &50s / 1.40h \\
    PNA &Transformer &RWSE-16 &0.1409 ± 0.0131 &7,551,217 &61s / 1.68h \\\midrule
    GatedGCN &Transformer &-- &0.1355 ± 0.0035 &6,155,089 &59s / 1.63h \\
    GatedGCN &Transformer &LapPE-8 &0.1201 ± 0.0003 &6,153,889 &63s / 1.76h \\
    GatedGCN &Transformer &SignNet\textsuperscript{MLP}-8 &0.1158 ± 0.0008 &6,217,013 &87s / 2.41h \\
    GatedGCN &Transformer &SignNet\textsuperscript{DeepSets}-21 &0.1144 ± 0.0002 &6,225,845 &146s / 4.05h \\
    GatedGCN &Transformer &PEG\textsuperscript{LapEig}-8 &0.1209 ± 0.0003 &6,162,390 &67s / 1.86h \\
    \bottomrule
    \end{tabular}
\end{table}

\begin{table}[ht]
    \caption{Ablation study on \textbf{MalNet-Tiny}. *Configuration required decreased batch size.
    %to fit the computation in a 32GB VRAM GPU.
    }
    \label{tab:abl_malnet}
    \centering
    % \footnotesize
    \fontsize{8.5pt}{8.5pt}\selectfont
    \begin{tabular}{ccccccc}\toprule
    \textbf{\method-MPNN} &\textbf{\method-GlobAttn} &\textbf{PE / SE type} &\textbf{Accuracy $\uparrow$} &\textbf{\# Param.} &\textbf{Epoch / Total} \\\midrule
    GatedGCN &Performer &-- &92.64 ± 0.78 &527,237 &46s / 1.90h \\\midrule
    GatedGCN &-- &-- &92.23 ± 0.65 &199,237 &6s / 0.25h \\
    GatedGCN &*Transformer &-- &93.50 ± 0.41 &282,437 &94s / 3.94h \\
    GatedGCN &BigBird &-- &92.34 ± 0.34 &324,357 &130s / 5.43h \\\midrule
    -- &Performer &-- &73.90 ± 0.58 &421,957 &41s / 1.73h \\
    GINE &Performer &-- &92.27 ± 0.60 &463,557 &46s / 1.92h \\
    PNA &Performer &-- &91.67 ± 0.70 &592,149 &47s / 1.97h \\\midrule
    GatedGCN &Performer &LapPE-10 &92.74 ± 0.45 &527,701 &47s / 1.91h \\
    GatedGCN &Performer &RWSE-16 &92.77 ± 0.31 &527,425 &46s / 1.90h \\
    GatedGCN &Performer &SignNet\textsuperscript{MLP}-10 &92.57 ± 0.40 &591,063 &65s / 2.72h \\
    GatedGCN &Performer &*SignNet\textsuperscript{DeepSets}-32 &93.13 ± 0.68 &602,085 &145s / 6.06h \\
    GatedGCN &Performer &PEG\textsuperscript{LapEig}-10 & 92.27 ± 0.29 &528,842 &48s / 1.98h\\
    \bottomrule
    \end{tabular}
\end{table}

\begin{table}[ht]
    \caption{Ablation study on \textbf{CIFAR10}.}
    \label{tab:abl_cifar10}
    \centering
    % \footnotesize
    \fontsize{8.5pt}{8.5pt}\selectfont
    \begin{tabular}{ccccccc}\toprule
    \textbf{\method-MPNN} &\textbf{\method-GlobAttn} &\textbf{PE / SE type} &\textbf{Accuracy $\uparrow$} &\textbf{\# Param.} &\textbf{Epoch / Total} \\\midrule
    GatedGCN &Transformer &LapPE-8 &72.305 ± 0.344 &112,726 &62s / 1.72h \\\midrule
    GatedGCN &-- &LapPE-8 &69.948 ± 0.499 &79,654 &43s / 1.18h \\
    GatedGCN &Performer &LapPE-8 &70.670 ± 0.338 &239,554 &77s / 2.14h \\
    GatedGCN &BigBird &LapPE-8 &70.480 ± 0.106 &129,418 &145s / 4h \\\midrule
    -- &Transformer &LapPE-8 &68.862 ± 1.138 &70,762 &40s / 1.11h \\
    GINE &Transformer &LapPE-8 &71.105 ± 0.655 &87,298 &51s / 1.42h \\
    PNA &Transformer &LapPE-8 &73.418 ± 0.165 &138,706 &59s / 1.65h \\\midrule
    GatedGCN &Transformer &-- &71.488 ± 0.187 &112,590 &61s / 1.69h \\
    GatedGCN &Transformer &RWSE-16 &71.958 ± 0.398 &112,798 &61s / 1.69h \\
    GatedGCN &Transformer &SignNet\textsuperscript{MLP}-8 &71.740 ± 0.569 &175,850 &116s / 3.21h \\
    GatedGCN &Transformer &SignNet\textsuperscript{DeepSets}-16 &72.368 ± 0.340 &186,558 &148s / 4.12h \\
    GatedGCN &Transformer &PEG\textsuperscript{LapEig}-8 &72.100 ± 0.460 &113,529 &67s / 1.87h \\
    \bottomrule
    \end{tabular}
\end{table}

%%%% LRGB ablations.
\begin{table}[ht]
    \caption{Ablation study on \textbf{PascalVOC-SP} of LRGB~\cite{dwivedi2022LRGB}. Shown is the mean~±~s.d.~of 4 runs.}
    \label{tab:abl_pascal}
    \centering
    % \footnotesize
    \fontsize{8.5pt}{8.5pt}\selectfont
    \begin{tabular}{ccccccc}\toprule
    \textbf{\method-MPNN} &\textbf{\method-GlobAttn} &\textbf{PE / SE type} &\textbf{F1 $\uparrow$} &\textbf{\# Param.} &\textbf{Epoch / Total} \\\midrule
    GatedGCN &Transformer &LapPE-10 &0.3736 ± 0.0158 &510,453 &17s / 1.46h \\\midrule
    GatedGCN &-- &LapPE-10 &0.3016 ± 0.0031 &361,461 &8s / 0.68h \\
    GatedGCN &Performer &LapPE-10 &0.3724 ± 0.0131 &1,148,277 &25s / 2.09h \\
    GatedGCN &BigBird &LapPE-10 &0.2762 ± 0.0069 &585,333 &42s / 3.46h \\\midrule
    -- &Transformer &LapPE-10 &0.2762 ± 0.0111 &322,677 &12s / 1.04h \\
    GINE &Transformer &LapPE-10 &0.3160 ± 0.0024 &397,173 &14s / 1.18h \\
    PNA &Transformer &LapPE-10 &0.3677 ± 0.0108 &625,029 &18s / 1.49h \\\midrule
    GatedGCN &Transformer &-- &0.3846 ± 0.0156 &510,069 &17s / 1.4h \\
    GatedGCN &Transformer &RWSE-16 &0.3659 ± 0.0031 &510,133 &17s / 1.45h \\
    GatedGCN &Transformer &SignNet\textsuperscript{MLP}-10 &0.3473 ± 0.0051 &573,869 &41s / 3.4h \\
    GatedGCN &Transformer &SignNet\textsuperscript{DeepSets}-48 &0.3668 ± 0.0080 &583,893 &50s / 2.8h \\
    GatedGCN &Transformer &PEG\textsuperscript{LapEig}-10 &\textbf{0.3956 ± 0.0084} &512,281 &19s / 1.6h \\
    \bottomrule
    \end{tabular}
\end{table}

\begin{table}[ht]
    \caption{Ablation study on \textbf{Peptides-func} of LRGB~\cite{dwivedi2022LRGB}. Shown is the mean~±~s.d.~of 4 runs.}
    \label{tab:abl_peptidesfunc}
    \centering
    % \footnotesize
    \fontsize{8.5pt}{8.5pt}\selectfont
    \begin{tabular}{ccccccc}\toprule
    \textbf{\method-MPNN} &\textbf{\method-GlobAttn} &\textbf{PE / SE type} &\textbf{AP $\uparrow$} &\textbf{\# Param.} &\textbf{Epoch / Total} \\\midrule
    GatedGCN &Transformer &LapPE-10 &0.6535 ± 0.0041 &504,362 &6s / 0.35h \\\midrule
    GatedGCN &-- &LapPE-10 &0.6159 ± 0.0048 &355,370 &3s / 0.16h \\
    GatedGCN &Performer &LapPE-10 &0.6475 ± 0.0056 &748,970 &11s / 0.61h \\
    GatedGCN &BigBird &LapPE-10 &0.5854 ± 0.0079 &579,242 &18s / 1.00h \\\midrule
    -- &Transformer &LapPE-10 &0.6333 ± 0.0040 &316,586 &5s / 0.29h \\
    GINE &Transformer &LapPE-10 &0.6464 ± 0.0077 &391,082 &6s / 0.31h \\
    PNA &Transformer &LapPE-10 &0.6560 ± 0.0058 &618,138 &6s / 0.35h \\\midrule
    GatedGCN &Transformer &-- &0.6214 ± 0.0326 &506,506 &6s / 0.33h \\
    GatedGCN &Transformer &RWSE-16 &0.6486 ± 0.0071 &503,418 &6s / 0.35h \\
    GatedGCN &Transformer &SignNet\textsuperscript{MLP}-10 &0.5840 ± 0.0140 &568,726 &41s / 3.39h \\
    GatedGCN &Transformer &SignNet\textsuperscript{DeepSets}-48 &0.6314 ± 0.0059 &577,802 &49s / 2.73h \\
    GatedGCN &Transformer &PEG\textsuperscript{LapEig}-10 &0.6461 ± 0.0047 &508,718 &19s / 1.60h \\
    \bottomrule
    \end{tabular}
\end{table}

%%%%%%%%%%%%%%%%%%%%%%%%%%%%%%%%%%%%%%%%%%%%%%%%%%%%%%%%%%%%%%%%%%%%%%%%%%%%%%%
\clearpage
\section{Theoretical results}\label{app:theory}
\setcounter{figure}{0}
\setcounter{table}{0}
%%%%%%%%%%%%%%%%%%%%%%%%%%%%%%%%%%%%%%%%%%%%%%%%%%%%%%%%%%%%%%%%%%%%%%%%%%%%%%%

%\section{Weisfeiler-Leman 1-WL test}
\subsection{Why do we need PE and SE?}
\label{app:WL}
% \db{TODO: Vijay :)} \vd{VD: I have elaborated the main paper's 3.2.2 in this appendix section, and included a conclusion statement at the end. Should we keep the statement? Feel free to revise the statement for more precision.}

In this section, we review the 1-Weisfeiler-Leman test \cite{weisfeiler1968reduction}, their equivalence with MPNNs and the limitations brought by this equivalent expressive power which eventually brings us to a statement that indicates the theoretical need of equipping MPNNs or GTs with either or a combination of local, relative or global PE/SE.

\paragraph{1-Weisfeiler-Leman test (1-WL).}
The 1-WL test is a node-coloring algorithm, in the hierarchy of Weisfeiler-Leman (WL) heuristics for graph isomorphism, \cite{weisfeiler1968reduction}, which iteratively updates the color of a node based on its 1-hop local neighborhood until an iteration when the node colors do not change successively. The final histogram of the node colors determine whether the algorithm outputs the two graphs to be `non-isomorphic' (when the histograms of 2 graphs are distinct) or `possibly isomorphic' (when the histograms of 2 graphs are same). Although, it is not a sufficient test for the graph isomorphism problem, the heuristic is simple to apply and has been popularly used in the literature recently to quantify the expressive power of MPNNs.

\paragraph{Expressive power of MPNNs.} Based on the equivalence of the aggregate and update functions of MPNNs with the hash function of the 1-WL test, it was shown that MPNNs are at most powerful as 1-WL \cite{xu2018how, morris2019}, which is now popularly understood in the literature. Graph Isomorphism Network \cite{xu2018how} was developed by aligning the injectivity of the aggregate and update functions of GIN with the injectivity of the 1-WL's hash function, which makes it a 1-WL powerful MPNN. In direct consequence, the power of the GIN is quantified as 1-WL expressive, \textit{i.e.}, if 1-WL outputs two graphs to be `non-isomorphic' then the GIN would output different feature vectors for the two graphs and conversely, if 1-WL outputs two graphs to be `possibly isomorphic', the feature embeddings of the two graphs would be the same. We refer the readers to \cite{xu2018how} for the details on this theoretical result.

\begin{figure}[ht]
\centering
\begin{subfigure}{.7\textwidth}
  \centering
  \includegraphics[height=11cm,keepaspectratio]{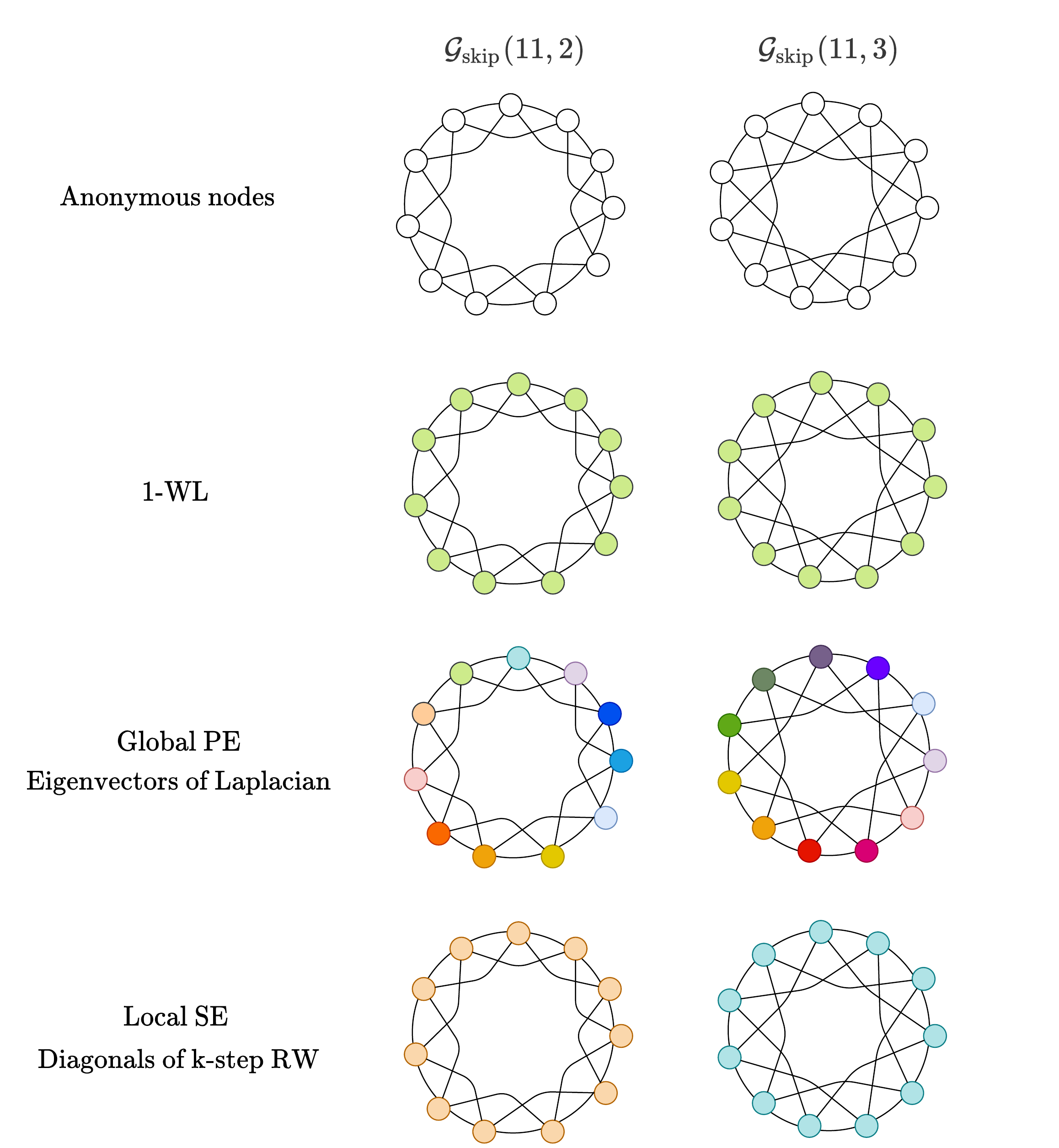}
  \subcaption{}
  \label{fig:pe_se_example:a}
\end{subfigure}%
\hfill%
\begin{subfigure}{.3\textwidth}
  \centering
  \includegraphics[height=11cm,keepaspectratio]{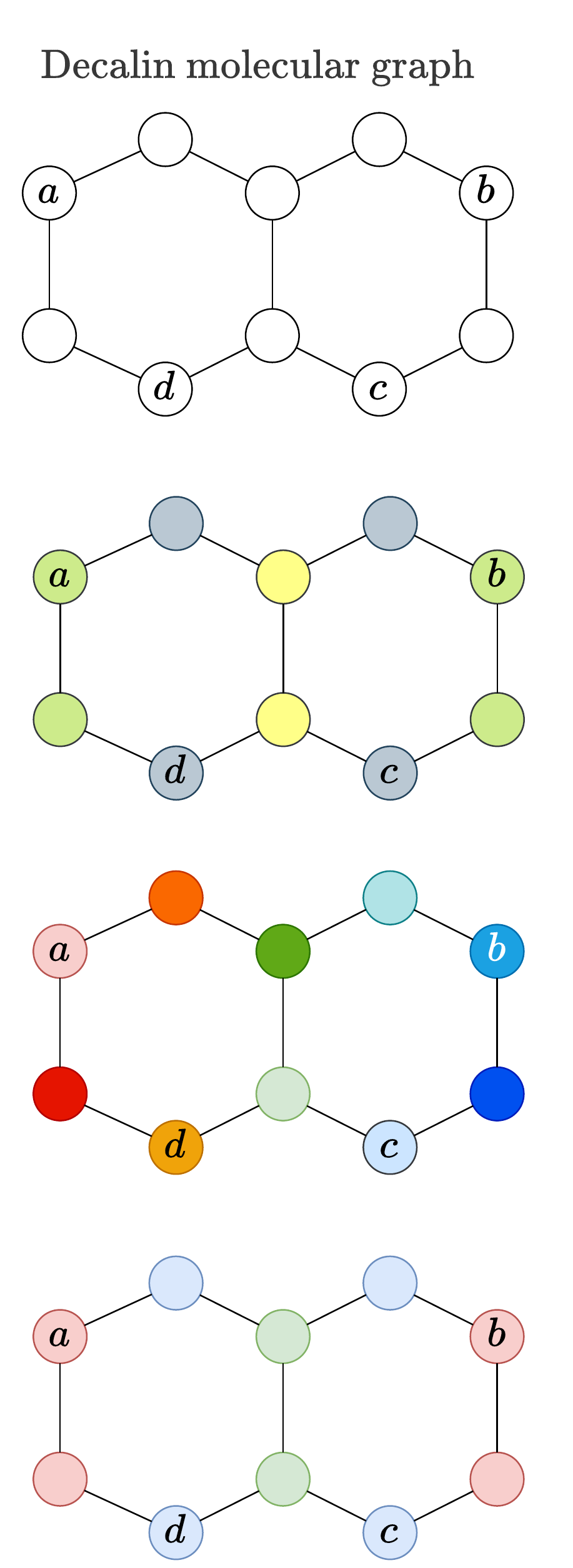}
  \subcaption{}
  \label{fig:pe_se_example:b}
\end{subfigure}
\caption{\textbf{First Row:} Example graphs with anonymous nodes, \textit{i.e.}, nodes do not have any distinguishing node features. (a) A pair of Circular Skip Link (CSL) graphs \cite{murphy2019relational} where the nodes have skip links of 2 and 3 respectively. (b) A Decalin molecular graph which has two rings of all Carbon atoms, thus with no distinguishing node features. \textbf{Second Row:} The nodes colored with the feature generated by 1-WL \cite{weisfeiler1968reduction, xu2018how, morris2019}.
  \textbf{Third Row:} The nodes colored with the feature generated by \textit{global} PE \cite{dwivedi2020benchmarking}. \textbf{Fourth Row:} The nodes colored with the feature generated by \textit{local} SE \cite{dwivedi2022LPE}. \textit{Note:} The colors depicted on nodes in the graphs represent a unique feature vector generated, for a given graph, from the corresponding PE/SE. Figure best visualized in color.}
  \label{fig:pe_se_example}
\end{figure}

Since the expressive power of MPNNs are at most 1-WL, it leads to a serious limitation in distinguishing a wide-variety of non-isomorphic graphs \cite{sato2020survey}. Note that numerous follow up works have proposed GNNs that are strictly powerful than 1-WL, often moving away from the message passing framework \cite{gilmer2017neural} on which MPNNs are based \cite{morris2019, chen2019equivalence, maron2019provably}. As higher-order GNNs are not within the scope of this section, we limit our discussion only to MPNNs, such as GINs, which makes them 1-WL powerful.
There are numerous examples on which MPNNs fail as a result \cite{sato2020survey}. Among such cases, we consider two examples tasks: the task to differentiate between two non-isomorphic Circular Skip Link (CSL) graphs, Figure \ref{fig:pe_se_example:a}, and the task to differentiate between two potential links, Figure \ref{fig:pe_se_example:b}. The nodes in these examples do not have discriminating node features.

% It is well known that standard MPNNs are as expressive as the 1-WL test, meaning that they fail to distinguish non-isomorphic graphs under a 1-hop aggregation. In this section, we argue that the selected \textit{local}, \textit{global} and \textit{relative} PE/SE allow MPNNs to become more expressive than the 1-WL by looking at the examples from  Figure~\ref{fig:pe_se_example} in the Appendix, thus making them fundamentally more expressive at distinguishing between nodes and graphs.

% \begin{figure}[ht]
%   \centering
%   \includegraphics[width=\textwidth]{./figures/pe_se_example_color.pdf}
%   \caption{\textbf{First Row:} Example graphs with anonymous nodes, \textit{i.e.}, nodes do not have any distinguishing node features. (i) A pair of Circular Skip Link (CSL) graphs \cite{murphy2019relational} where the nodes have skip links of 2 and 3 respectively. (ii) A Decalin molecular graph which has two rings of all Carbon atoms, thus with no distinguishing node features. \textbf{Second Row:} The nodes colored with the feature generated by 1-WL \cite{weisfeiler1968reduction, xu2018how, morris2019}.
%   \textbf{Third Row:} The nodes colored with the feature generated by Global PE \cite{dwivedi2020benchmarking}. \textbf{Fourth Row:} The nodes colored with the feature generated by Local SE \cite{dwivedi2022LPE}. \textit{Note:} The colors depicted on nodes in the graphs represent a unique feature vector generated, for a given graph, from the corresponding PE.}
%   \label{fig:pe_se_example}
% \end{figure}

% \begin{figure}[ht]
%   \centering
%   \includegraphics[width=0.75\textwidth]{./figures/pe_se_example.pdf}
%   \caption{Example graphs with anonymous nodes, \textit{i.e.}, nodes do not have any distinguishing node features. (i) A pair of Circular Skip Link (CSL) graphs \cite{murphy2019relational} where the nodes have skip links of 2 and 3 respectively. (ii) A Decalin molecular graph which has two rings of all Carbon atoms, thus with no distinguishing node features.
%   }
%   \label{fig:pe_se_example}
% \end{figure}

\paragraph{The CSL graph, Figure \ref{fig:pe_se_example:a}.}
In the CSL graph-pair \cite{murphy2019relational},
% Figure~\ref{fig:pe_se_example:a}, 
the two graphs $\mathcal{G}_{\textrm{\tiny skip}}(11,2)$ and $\mathcal{G}_{\textrm{\tiny skip}}(11,3)$ differ in the length of skip-link of a node and are hence non-isomorphic. Since the 1-WL algorithm produces the same color for all the nodes in both graphs, MPNNs will generate similar node colors. See the colors generated by 1-WL and MPNN in the second row of Figure~\ref{fig:pe_se_example:a}. 
However, the use of a \textit{global} PE (eg. Laplacian PE \cite{dwivedi2020benchmarking}) assigns each node a unique color, as depicted in the third row. Consequently, the feature embeddings of the two graphs which are the hash function outputs of the collection of node colors are different, thus making the task to distinguish the graphs successful.
Similarly, the use of a \textit{local} SE (e.g. diagonals of $m$-steps random walk) allows the coloring of the nodes of the 2 graphs to be different \cite{dwivedi2022LPE} since it captures the difference of the skip links of the two graphs successfully \cite{loukas2020graph}. See the fourth row where the local SE based colors are depicted on the nodes. Therefore, either of the specific \textit{local} SE or \textit{global} PE can help distinguish the two graphs which cannot be learnt by 1-WL or MPNNs.

\paragraph{The Decalin molecular graph, Figure \ref{fig:pe_se_example:b}.}
In the Decalin graph,
% presented in Figure \ref{fig:pe_se_example:b},
the node $a$ is isomorphic to node $b$, and so is the node $c$ to node $d$. A 1-WL coloring of the nodes, and equivalently MPNN, would generate one color for the node $a,b$ and another color for $c,d$, see the second row in Figure \ref{fig:pe_se_example:b}. If that task is to identify a potential link between the node-sets $(a,d)$ and $(b,d)$, the combination of the node colors of the node-sets will produce the same embedding for the two links, thus making the 1-WL or MPNNs based coloring unsuitable to certain tasks \cite{zhang2021labeling}.
A similar observation also follows for the node coloring based on the aforementioned \textit{local} SE \cite{dwivedi2022LPE}, which is illustrated in the fourth row in Figure \ref{fig:pe_se_example:b}. However, using a distance-based \textit{relative} PE on the edges or an eigenvector-based \textit{global} PE would successfully differentiate the embeddings of the two links.
Therefore, the \textit{relative} PE or the \textit{global} PE which can help to distinguish between the two links cannot be learnt by 1-WL or MPNNs.

% It henceforth justifies the theoretical need to encode PE and SE at any or all of the 3 levels discussed to make the MPNN strictly powerful than 1-WL.

We can then conclude the following statement based on the above discussion which provides a theoretical basis for the need of PE and SE, as the PE and SE can be directly supplying essential information for the task:

\textbf{Proposition 1.} \textit{Assuming no modification applied to MPNNs for a learning task, there exists Positional Encodings (PE) and Structural Encoding (SE) which MPNNs are not guaranteed to learn.}

\newpage
\subsection{Preserving edge information in the self-attention layer}
\label{app:edge-in-attention}

In this section, we argue that an MPNN layer is able to propagate the information from edges to nodes such that, when computing the attention between nodes, the global Attention (Transformer) layer can infer whether two nodes are connected and what are the edge features between them.

Suppose an MPNN with the sum aggregator, with the update function as given below:
\begin{equation}
    h_u^{l+1} = \sum_{v\in\mathcal{N}_u}{f(h_u^l, h_v^l, e_{uv})},
\end{equation}
where $f$ is a learned function, e.g., an MLP; $u$ is the index of a central node whose neighborhood is being aggregated; $v$ is the index of a neighbor of $u$; $h^l_u$ the node features at layer $l$ for node $u$, and $e_{uv}$ the edge features between nodes $u$ and $v$.

We know from the Lemma 5 of \citet{xu2018how} that the sum over a countable multiset is universal, meaning it can map a unique multiset to any possible function. Let's assume that $h_u$ is unique and countable for every node $u$, which can be accomplised using all the Laplacian eigenvectors as PE. Then, there exist a function $f$ such that an encoding $\mu_{uv}$ that respects the following characteristics is propagated to the nodes: (i) unique for the triplet $\{h_u, h_v, e_{uv}\}$, (ii) invariant to the permutation of $u$ and $v$, (iii) contains the information of $e_{ij}$, (iv) all information of $\mu_{uv}$ is preserved after the $\sum$.

Hence, an Attention layer that follows the message-passing is able to infer whether two nodes are connected since both nodes will contain the unique identifier $\mu_{uv}$, and will also be able to infer the edge features from it.

An example of such function $\mu_{uv}$ is the tensor product $\otimes$ of a one-hot encoding unique for each edge $o_{uv}$ and the edge features $e_{uv}$. For example, if $e_{uv} = [e_1, e_2, e_3]$ and the edge is represented with $o_{uv} = [0, 1, 0, 0]$, then $\mu_{uv} = o_{uv} \otimes e_{uv} = [0, 0, 0, e_1, e_2, e_3, 0, 0, 0, 0, 0,0]$ satisfies all the above conditions. Although this function requires an exponential increase in the hidden dimension, this is also the case for the Lemma 5 in \citet{xu2018how}.

%%%%%%%%%%%%%%%%%%%%%%%%%%%%%%%%%%%%%%%%%%%%%%%%%%%%%%%%%%%%%%%%%%%%%%%%%%%%%%%
\clearpage
\section{\method schematics}
\label{app:GPS-layer}
\setcounter{figure}{0}
\setcounter{table}{0}
%%%%%%%%%%%%%%%%%%%%%%%%%%%%%%%%%%%%%%%%%%%%%%%%%%%%%%%%%%%%%%%%%%%%%%%%%%%%%%%

\subsection{GPS layer}

\begin{figure}[ht]
  \centering
  \includegraphics[width=0.9\textwidth]{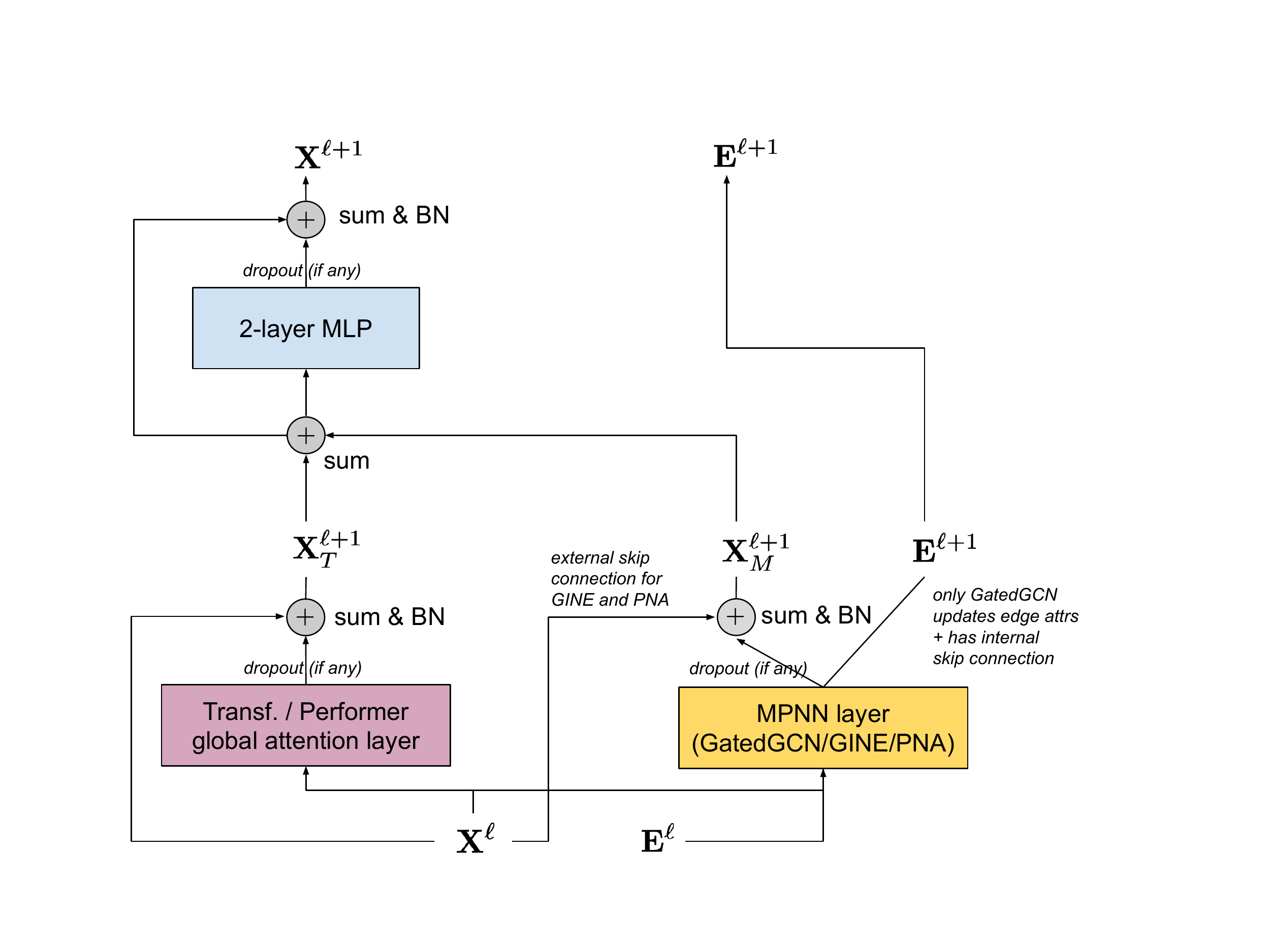}
  \caption{Modular \method layer that combines local MPNN and global attention blocks. Local MPNN encodes real edge features into the node-level hidden representations, while global attention mechanism can implicitly make use of this information together with PE/SE to infer relation between two nodes without explicit edge features. After each functional block (an MPNN layer, a global attention layer, an MLP) we apply residual connections followed by batch normalization (BN)~\cite{ioffe2015batchnorm}. In the 2-layer MLP block we use ReLU activations and its inner hidden dimension is twice the layer-input feature dimensionality $d_\ell$. Note, similarly to Transformer, the input and output dimensionality of the \method-layer as a whole is the same.
  }
  \label{fig:gps_layer}
\end{figure}

\newcommand{\drp}{\texttt{Dropout}\xspace}
\newcommand{\bn}{\texttt{BatchNorm}\xspace}
\myparagraph{\method layer equations.} In Section~\ref{sec:gps_layer} of the main text we provide a simplify definition of the \method computational layer for clarity, here we additionally list the precise application of skip connections, dropout, and batch normalization with learnable affine parameters:
\begin{eqnarray}
    \X^{\ell+1}, \E^{\ell+1} &=& \texttt{GPS}^{\ell} \left( \X^{\ell}, \E^{\ell}, \A \right)\\
    \textrm{computed as} \ \ \ \ 
    \hat\X^{\ell+1}_M, \ \E^{\ell+1} &=& \mpnn_e^{\ell} \left(\X^{\ell}, \E^{\ell}, \A \right),\\
    \hat\X^{\ell+1}_T &=&  \attn^{\ell} \left(\X^{\ell}  \right),\\
    \X^{\ell+1}_M   &=& \bn \left( \drp \left( \hat\X^{\ell+1}_M \right) + \X^{\ell} \right), \\
    \X^{\ell+1}_T   &=& \bn \left( \drp \left( \hat\X^{\ell+1}_T \right) + \X^{\ell} \right), \\
    \X^{\ell+1}     &=& \texttt{MLP}^{\ell}\left(\X^{\ell+1}_M + \X^{\ell+1}_T\right)
\end{eqnarray}

\newpage
\subsection{GPS algorithm}

\newcommand{\nodeenc}{\texttt{NodeEncoder}\xspace}
\newcommand{\edgeenc}{\texttt{EdgeEncoder}\xspace}

\begin{algorithm}
\caption{Algorithm for an $L$ layer GPS network.}

\textbf{Input:} Graph $\mathcal{G} = (\mathcal{V}, \mathcal{E})$ with $N$ nodes and $E$ edges; 
Adjacency matrix $\A \in \mathbb{R}^{N \times N}$; 
Node features $\X \in \mathbb{R}^{N \times D_\textrm{\tiny  node} }$; 
Edge features $\E \in \mathbb{R}^{E \times D_\textrm{\tiny edge} }$;
Local message passing model instance $\mpnn_e$;
Global attention model instance $\attn$; 
Positional Encoding function $F_\textsc{pe}$;
Structural Encoding function $F_\textsc{se}$;
Layer $\ell \in [0,L-1]$.\\ 
\textbf{Output:} Node representations $\X^L \in \mathbb{R}^{N \times D}$ and edge representations $\E^L \in \mathbb{R}^{E \times D}$, that can downstream be composed with appropriate \emph{prediction head} for graph, node, or edge -level prediction.
\begin{enumerate}
    \item $\mathbf{P}_\textrm{\tiny node}, \mathbf{P}_\textrm{\tiny edge}, \mathbf{S}_\textrm{\tiny node}, \mathbf{S}_\textrm{\tiny edge} \ \gets \ \emptyset$
    \item if $F_\textsc{pe}$ is relative then $\mathbf{P}_\textrm{\tiny edge} \ \gets F_\textsc{pe}(\mathcal{G}) \ \in \mathbb{R}^{E \times D_\textsc{pe}}$ \ else \ $\mathbf{P}_\textrm{\tiny node} \ \gets F_\textsc{pe}(\mathcal{G}) \ \in \mathbb{R}^{N \times D_\textsc{pe}} $
    \item if $F_\textsc{se}$ is relative then $\mathbf{S}_\textrm{\tiny edge} \ \gets F_\textsc{se}(\mathcal{G}) \ \in \mathbb{R}^{E \times D_\textsc{se}}$ \ else \ $\mathbf{S}_\textrm{\tiny node} \ \gets F_\textsc{se}(\mathcal{G}) \ \in \mathbb{R}^{N \times D_\textsc{se}} $
    \item $\X^{0} \ \gets \ \bigoplus_\textrm{\tiny node} \bigl( \nodeenc \left(\X\right), \mathbf{P}_\textrm{\tiny node}, \mathbf{S}_\textrm{\tiny node} \bigr) \ \in \mathbb{R}^{N \times D}$ 
    \item $\E^{0} \ \gets \ \bigoplus_\textrm{\tiny edge} \bigl( \edgeenc \left(\E\right), \mathbf{P}_\textrm{\tiny edge}, \mathbf{S}_\textrm{\tiny edge} \bigr) \ \in \mathbb{R}^{E \times D}$ 
    \item for $\ell = 0, 1, \cdots , L-1 $
    \begin{enumerate}
        \item $\hat\X^{\ell+1}_M, \ \E^{\ell+1} \gets \mpnn_e^{\ell} \left(\X^{\ell}, \E^{\ell}, \A \right)$
        \item $\hat\X^{\ell+1}_T \gets  \attn^{\ell} \left(\X^{\ell}  \right)$
        \item $\X^{\ell+1}_M   \gets \bn \left( \drp \left( \hat\X^{\ell+1}_M \right) + \X^{\ell} \right)$
        \item $\X^{\ell+1}_T   \gets \bn \left( \drp \left( \hat\X^{\ell+1}_T \right) + \X^{\ell} \right)$
        \item $\X^{\ell+1}     \gets \texttt{MLP}^{\ell}\left(\X^{\ell+1}_M + \X^{\ell+1}_T\right)$
    \end{enumerate}
    \item return $\X^{L} \in \mathbb{R}^{N \times D}$ and $\E^{L} \in \mathbb{R}^{E \times D}$
\end{enumerate}
\label{algo:gps}
\end{algorithm}
where $\bigoplus$ denotes an operator for combining the input node or edge features with their respective positional and/or structural encoding, in practice this is a concatenation operator which can be changed to sum or other operators; $\nodeenc$ and $\edgeenc$ are dataset-specific initial node and edge feature encoders potentially with learnable parameters; $\mpnn_e$ and $\attn$ have their corresponding learnable parameters at each layer $\ell$; $\hat\X^{\ell+1}_M$ and $\hat\X^{\ell+1}_T$ denote the intermediate node representations given by the local message passing module and the global attention module respectively; and $\texttt{MLP}^{\ell}$ is a multi layer perceptron module with its own learnable parameters that combines the intermediate $\X^{\ell+1}_M$ and $\X^{\ell+1}_T$. Note that a relative $F_\textsc{pe}$ or $F_\textsc{se}$ produces PE or SE for each edge which are thence handled accordingly in lines 2 and 3 in Algorithm \ref{algo:gps}.

\end{document}